%% file: main.tex
\documentclass[10pt,twocolumn,letterpaper]{article}

\usepackage{amsthm}
\newtheorem{theorem}{Theorem}
\usepackage{threeparttable}
\usepackage{multirow}
\usepackage{algorithm}
\usepackage{algorithmic} 

\usepackage[subtle]{savetrees}
\usepackage{lineno}

\usepackage[pagenumbers]{cvpr} 


\definecolor{cvprblue}{rgb}{0.21,0.49,0.74}
\usepackage[pagebackref,breaklinks,colorlinks,allcolors=cvprblue]{hyperref}

\usepackage{xcolor}
\definecolor{darkergreen}{rgb}{0.0, 0.5, 0.0} 

\begin{document}

\def\paperID{9273} 
\def\confName{CVPR}
\def\confYear{2025}

\title{Improve Representation for Imbalanced Regression \\ through Geometric Constraints}


\author{
  Zijian~Dong$^{1, }$\thanks{\emph{Equal contribution}} \ \ Yilei Wu$^{1, }$\footnotemark[1] \ \ Chongyao Chen$^{2, }$\footnotemark[1] \ \ Yingtian Zou$^{1}$ \ \ Yichi Zhang$^{1}$ \ \ Juan Helen Zhou$^{1, }$\thanks{\emph{Corresponding author}} \\[8pt]
  $^1$National University of Singapore, \ \ $^2$ Duke University \\[8pt]
  \texttt{\{zijian.dong, yilei.wu\}@u.nus.edu, helen.zhou@nus.edu.sg}
  \\
}

\maketitle

\begin{abstract}
In representation learning, uniformity refers to the uniform feature distribution in the latent space (i.e., unit hypersphere). Previous work has shown that improving uniformity contributes to the learning of under-represented classes. However, most of the previous work focused on classification; the representation space of imbalanced regression remains unexplored. Classification-based methods are not suitable for regression tasks because they cluster features into distinct groups without considering the continuous and ordered nature essential for regression. In a geometric aspect, we uniquely focus on ensuring uniformity in the latent space for imbalanced regression through two key losses: \textbf{enveloping} and \textbf{homogeneity}. The enveloping loss encourages the induced trace to uniformly occupy the surface of a hypersphere, while the homogeneity loss ensures smoothness, with representations evenly spaced at consistent intervals. Our method integrates these geometric principles into the data representations via a \textbf{S}urrogate-driven \textbf{R}epresentation \textbf{L}earning 
(\textbf{SRL}) framework. Experiments with real-world regression and operator learning tasks highlight the importance of uniformity in imbalanced regression and validate the efficacy of our geometry-based loss functions. Code is available \hypersetup{urlcolor=[rgb]{0.96,0.43,0.78}}\href{https://github.com/yilei-wu/imbalanced-regression}{here}.
\end{abstract}

\section{Introduction}
\label{sec:intro}

Imbalanced datasets are ubiquitous across various domains, including image recognition \cite{zhang2023deep}, semantic segmentation \cite{zhong2023understanding}, and regression \cite{yang2021delving}. Previous studies have demonstrated the significance of uniform or balanced distribution of class representations for effective imbalanced classification \cite{wang2020understanding,kang2019decoupling,yin2019feature,kang2020exploring,cui2021parametric,li2022targeted,zhu2022balanced} and imbalanced semantic segmentation \cite{zhong2023understanding}. In classification tasks, these representations typically form distinct clusters. However, in the context of regression, representations are expected to be continuous and ordered \cite{zha2023rank,wu2023mixup}, rendering the methods used for quantifying and analyzing uniformity in classification inapplicable. While the issue of deep imbalanced regression (DIR) has received considerable attention, the focus has predominantly been on training unbiased regressors, rather than on the aspect of representation learning \cite{yang2021delving,steininger2021density,gong2022ranksim,ren2022balanced,keramati2023conr}. Among the methods that do explore representation learning \cite{zha2023rank,wu2023mixup}, the emphasis is typically on understanding the relationship between the label space and the feature space (the representations themselves should be continuous and ordered). However, a critical aspect that remains under-explored is the interaction between data representations and the entire feature space. Specifically, how these representations distribute within the full scope of the feature space has not been examined.

\begin{figure}[t]
    \centering
    \includegraphics[width=0.8\linewidth]{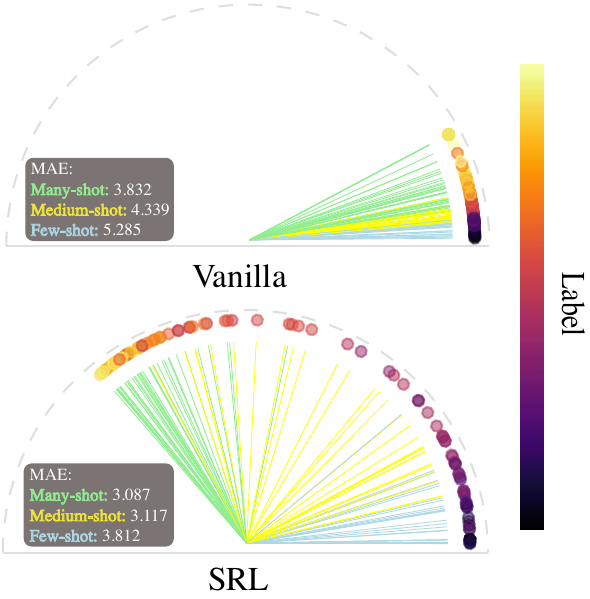}
    \caption{2D feature space of vanilla baseline and ours from UCI-Airfoil \cite{asuncion2007uci}. The vanilla feature space lacks uniformity and is dominated by samples from the Many-shot region. In contrast, our approach achieves a more uniform distribution over the feature space, improving the performance, especially in the Medium and Few-shot regions. (For visualization purposes, we curated the dataset to ensure equal partitions across the three regions.)}
    \label{fig1}
\end{figure}

\begin{figure*}[ht]
    \centering
    \includegraphics[width=2\columnwidth]{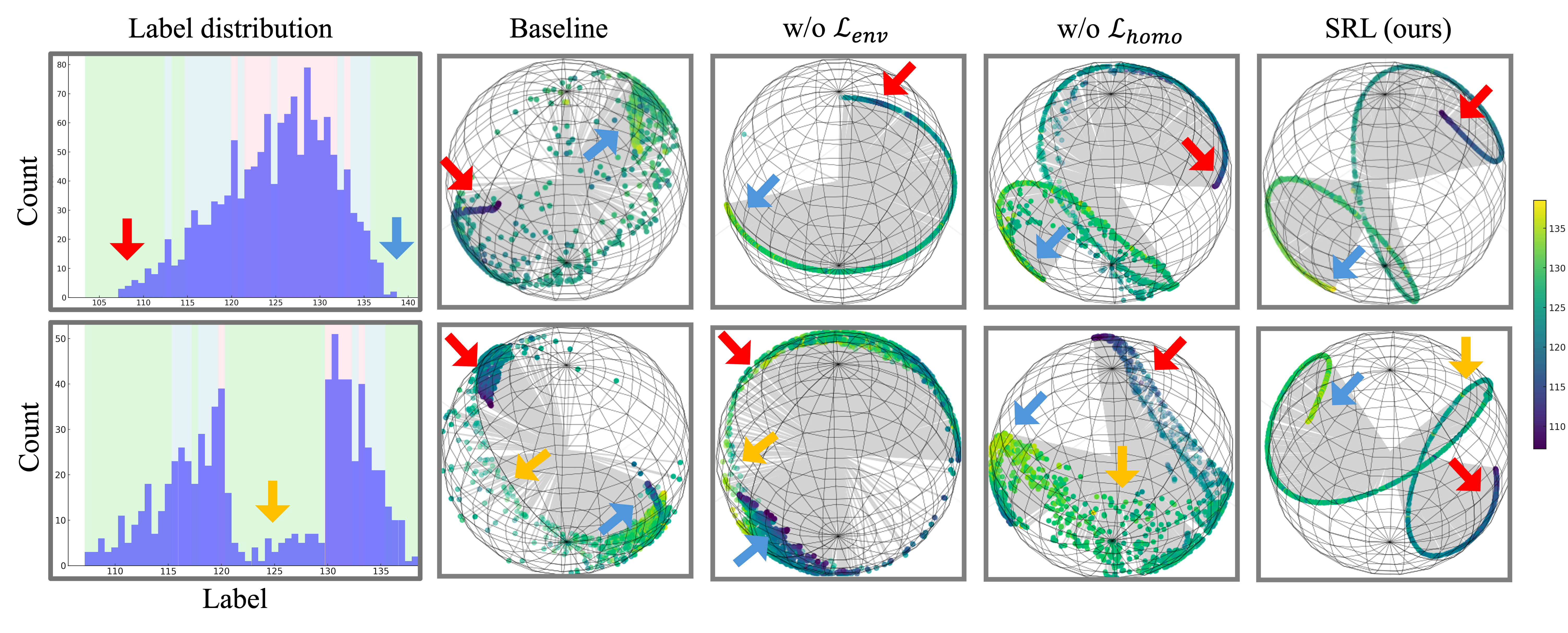}
    \caption{t-SNE visualization \citep{van2008visualizing} of feature comparison. The first row corresponds to the original UCI Airfoil Dataset \citep{asuncion2007uci}, while the second row corresponds to its curated version, with an additional few-shot region in the middle of the label range. Colored arrows point to the few-shot regions and their corresponding positions in the feature distributions. We evaluate feature distributions using: MSE Loss (Baseline), SRL without uniformity loss (w/o $\mathcal{L}_{\text{env}}$), SRL without homogeneity loss (w/o $\mathcal{L}_{\text{homo}}$), and complete SRL (ours). The baseline leads to feature collapse to many-shot regions and inadequate distinction of few-shot samples. In w/o $\mathcal{L}_{\text{env}}$, features collapse into a trivial shape, not fully utilizing the feature space. In w/o $\mathcal{L}_{\text{homo}}$, features spread out along the trace. Different from the previous ones, our SRL uniformly and smoothly ``fills" the feature space.} 
    \label{fig2}
    \vspace{-0.4cm}
\end{figure*}

Uniformity in classification refers to how effectively different clusters or centroids occupy the feature space, essentially partitioning it among various classes. In regression, here we define the term ``latent trace" as the pathway that the representations follow, delineating the transition from the minimum to the maximum label values. In this paper, we aim to evaluate \emph{\textbf{how well a latent trace occupies the feature space?}} To quantify this, we approximate a tubular neighborhood around the latent trace and measure its volume relative to the entire feature space. This method gauges the effectiveness of the trace in ``enveloping" the hypersphere, and we call it \textbf{enveloping loss}. This loss ensures that the trace shape fills the surface of the hypersphere to facilitate uniformity. In parallel, it is equally important that the points (\emph{i.e.}, individual data point representations) are evenly distributed along the trace. To address uniform distribution along the trace as well as smoothness, we have developed a \textbf{homogeneity loss}. This loss is computed based on the arc length of the trace, allowing us to effectively measure and promote an even and smooth distribution of points.

We model the uniformity in regression in two aspects: the induced trace aims to fully occupy the surface of the hypersphere (Enveloping), exhibiting smoothness with representations spaced at uniform intervals (Homogeneity). The two losses we introduce act as geometric constraints on a latent trace, implying they should not be applied to a set of representations from a single mini-batch. This is because a single batch likely does not encompass the full range of labels. To address this, we have developed a \textbf{S}urrogate-driven \textbf{R}epresentation \textbf{L}earning  (\textbf{SRL}) scheme. It involves averaging representations of the same bins within a mini-batch to form centroids and ``re-filling" missing bins by taking corresponding centroids from the previous epoch. This process results in a surrogate containing centroids for all bins, enabling the effective application of geometric loss across the complete label range. Furthermore, we introduce \emph{Imbalanced Operator Learning (IOL)} as a new DIR benchmark for training models on imbalanced domain locations in function space mapping. In summary, our main contributions are four-fold:

\begin{itemize}
  \item \textbf{Geometric Losses for Uniformity in deep imbalanced regression (DIR).} To the best of our knowledge, this work is the first to study representation learning in DIR. We introduce two novel loss functions, enveloping loss and homogeneity loss, to ensure uniform feature distribution for DIR.
  \item \textbf{SRL Framework.} A new framework is proposed that incorporates these geometric principles into data representations.
  \item \textbf{Imbalanced Operator Learning (IOL).} For the first time, we pioneer the task of operator learning within the realm of deep imbalanced regression, introducing an innovative task: Imbalanced Operator Learning (IOL). 
  \item \textbf{Extensive Experiments.} The effectiveness of the proposed method is validated through experiments involving real-world regression and operator learning, on five datasets: AgeDB-DIR, IMDB-WIKI-DIR, STS-B-DIR, and two newly created DIR benchmarks, UCI-DIR and OL-DIR.
\end{itemize}

\section{Related Work}
\label{sec:re}

\textbf{Uniformity in imbalanced classification.} \citet{wang2020understanding} identifies that uniformity is one of the key properties in contrastive representation learning. To promote uniformity in representation space for imbalanced classification, a variety of training strategies have been proposed. \citet{kang2019decoupling} decouples the training into a two-stage training of representation learning and classification. \citet{yin2019feature} designs a transfer learning framework for imbalanced face recognition. \citet{kang2020exploring} combines supervised method and contrastive learning to learn a discriminative and balanced feature space. PaCo \cite{cui2021parametric} and TSC \cite{li2022targeted} learn a set of class-wise balanced centers. BCL \cite{zhu2022balanced} balances the gradient distribution of negative classes and data distribution in mini-batch. Recent study suggests that sample-level uniform distribution may not effectively address imbalanced classification, advocating for category-level uniformity instead \cite{zhou2024combating,assran2022hidden}. Though progress has been made in this field, challenges persist in adapting the approach of modeling uniformity from classification to regression.

\noindent\textbf{Deep imbalanced regression.} With imbalanced regression data, effective learning in a continuous label space focuses on modeling the relationships among labels in the feature space \cite{yang2021delving}. Label Distribution Smoothing (LDS) \cite{yang2021delving} and DenseLoss \cite{steininger2021density} apply a Gaussian kernel to the observed label density, leading to an estimated label density distribution. Feature distribution smoothing (FDS) \cite{yang2021delving} generalizes the application of kernel smoothing from label to feature space. Ranksim \cite{gong2022ranksim} aims to leverage both local and global dependencies in data by aligning the order of similarities between labels and features. Balanced MSE \cite{ren2022balanced} addresses the issue of imbalance in Mean Squared Error (MSE) calculations, ensuring a more balanced distribution of predictions. VIR \cite{wang2023variational} provides uncertainty for imbalanced regression. ConR \cite{keramati2023conr} regularizes contrastive learning in regression by modeling global and local label similarities in feature space. RNC \cite{zha2023rank} and SupReMix \cite{wu2023mixup} learn a continuous and ordered representation for regression through supervised contrastive learning. How imbalanced regression representations leverage the feature space remains under-explored.

\section{Method}
\label{sec:method}

In the field of representation learning for classification, the concept of uniformity is pivotal for maximizing the use of the feature space \cite{wang2020understanding,kang2019decoupling,yin2019feature,kang2020exploring,cui2021parametric,li2022targeted,zhu2022balanced}. This idea is based on the principle of ensuring that features from different classes are not only distinctly separated but also evenly distributed in the latent space. This uniform distribution of class centroids fosters a clear and effective decision boundary, leading to more accurate classification. However, in regression, where we deal with continuous, ordered trace \cite{zha2023rank,wu2023mixup} rather than discrete clusters, \emph{\textbf{the concept of uniformity is not only more complex but essentially remains undefined}}.

We draw an analogy to the process of winding yarn around a ball. In this analogy, the yarn represents the latent trace, and the ball symbolizes the entirety of the available feature space. Just as the yarn must be evenly distributed across the ball's surface to effectively cover it (without any crossing), the latent trace should strive to occupy the hypersphere of the latent space uniformly. This ensures that the model leverages the available feature space to its fullest extent, enhancing the model's ability to capture the variability inherent in the data. 

Furthermore, the latent trace should be smooth and continuous, akin to the even stretching of yarn, rather than loose and disjointed. This smoothness ensures a consistent and predictable model behavior, which is crucial for the accurate prediction and interpretation of results.

We outline our method in this section. Firstly, we establish the fundamental notations and preliminaries (Section \ref{sec3.1}). Following this, we delve into the concept and definition of our enveloping loss (Section \ref{sec3.2}) and homogeneity loss (Section \ref{sec3.3}). Finally, we present our Surrogate-driven Representation Learning (SRL) framework, which incorporates the geometric constraints from the global image of the representations into the local range (Section \ref{sec3.4}). Refer to Supplementary Material \ref{b} for the pseudo code of our method.

\subsection{Preliminaries}
\label{sec3.1}

A regression dataset is composed of pairs $(\mathbf{x}_i,y_i)$, where $\mathbf{x}_i$ represents the input and $y_i$ is the corresponding continuous-value target. Denote $\mathbf{z}_i = f(\mathbf{x}_i)$ as the feature representation of $\mathbf{x}_i$, generated by a neural network $f(\cdot)$. The feature representation is normalized so that $\left\|\mathbf{z}_i\right\|=1$ for all $i$. Suppose the dataset consists of $K$ unique bins \footnote{The binning process (only for geometric loss calculations) facilitates surrogate formulation by using discrete centroids to approximate a continuous trace. Binning is unnecessary for most DIR datasets since they come pre-binned. The performed regression tasks follow the original dataset settings.}, we define a \emph{surrogate} as a set of centroids $\mathbf c_k$, where each represents a distinct bin. These centroids are computed by averaging the representations $\mathbf{z}$ sharing the same bin and they are normalized to $\left\|\mathbf{c}_k\right\|=1$. Let $l$ be a path: $l:[y_{\mathrm{min}},y_{\mathrm{max}}] \mapsto \mathbb{R}^n$ with $\left\|l(y)\right\|=1$, such that $l( y_k)  = \mathbf c_k$. The path $l$ is a continuous curve extended from the discrete dataset that lies on a submanifold of $\mathbb{R}^n$.

\begin{figure}[t]
    \centering
    \includegraphics[width=\linewidth]{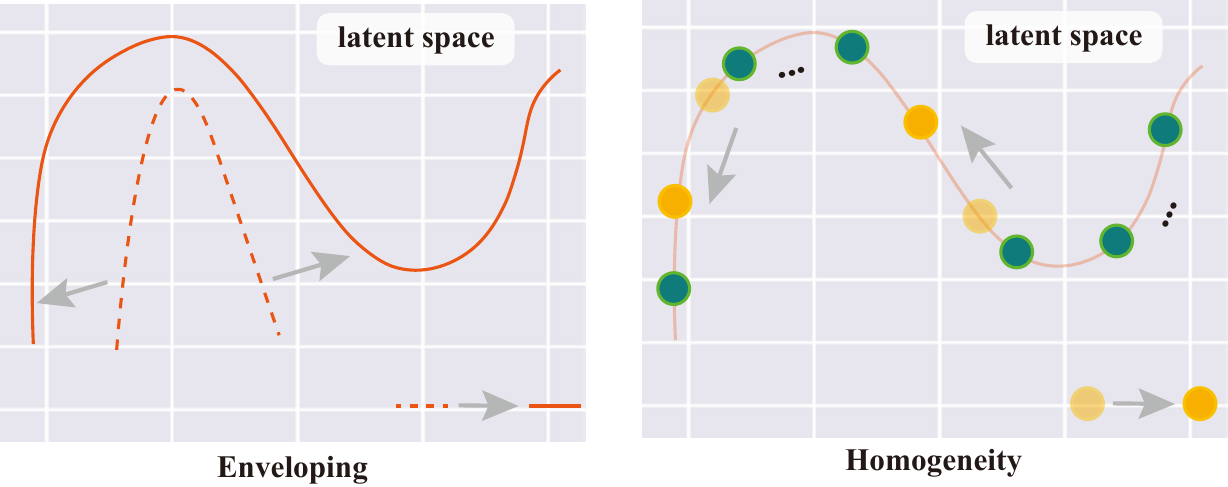}
    \caption{2D schematic overview of two geometric losses. The arrow indicates the improvement of the loss function. Enveloping loss encourages the representations to fill the latent space, and homogeneity loss encourages the smoothness and even distribution of the representations along the trace.}
    \label{fig3}
\end{figure}

\subsection{Enveloping}
\label{sec3.2}

To maximize the use of the feature space, it is crucial for the ordered and continuous trace of regression representations to fill the entire unit hypersphere as much as possible. This is by analogy with wrapping yarn with a certain length around a ball (without any crossing), aiming to cover as much surface area as possible. 

The trace of regression representations lying on a submanifold of $\mathbb{R}^n$ has a negligible hypervolume, which makes it challenging to assess its relationship with the entire hypersphere. To address this challenge, we extend the ``line" into a tubular neighborhood. This expansion allows us to introduce the concept of \emph{enveloping loss}. Our objective with this loss function is to maximize the hypervolume of the tubular neighborhood in proportion to the total hypervolume of the hypersphere.

Denote the set of all unit vectors in $\mathbb{R}^n$ as $\mathcal{U}$. Given $\epsilon \in (0,1)$, define tubular neighborhood $T(l,\epsilon)$ of $l$ as:
\begin{equation}
    T(l,\epsilon)=\{ \mathbf{z} \in \mathcal{U} \ | \  \mathbf{t}\cdot\mathbf{z} > \epsilon \ \text{for some} \ \mathbf{t} \in \mathrm{Im}(l)\}
\end{equation}
where for a function $f:A\rightarrow B$, the image is defined as $\mathrm{Im}(f) := \{f(x),x\in A\}$.

Then our enveloping loss is defined as:
\begin{equation}
    \mathcal{L}_{\text{env}} = -\frac{\text{vol}(T(l,\epsilon))}{\text{vol}(\mathcal{U})}
\end{equation}
where vol($\cdot$) returns the hypervolume of its input in the induced measure from the Euclidean space. 

In practical scenarios, the trace is composed of discrete representations, which complicates the direct computation of the tubular neighborhood's hypervolume. To navigate this challenge, we propose a continuous-to-discrete strategy. We first generate $N$ points that are uniformly distributed across the hypersphere. We then determine the fraction of these points that fall within the neighbourhood $\epsilon$. This fraction effectively approximates the proportion of the hypersphere covered by the tubular neighborhood with a sufficiently large $N$. To adapt $\mathcal{L}_{\text{env}}$ to discrete datasets, we re-formalize our optimization objective as:

\begin{equation}
\max \lim_{N\rightarrow\infty}\frac{P(N)}{N}
\end{equation}
where
\begin{equation}
    P(N):=|\{\mathbf{p}_i \ | \ \max_{y}\{\mathbf{p}_i\cdot l(y)\}>\epsilon, i \in [N]\}|
\label{eq4}
\end{equation}

assuming for each $N>0$, we can choose $N$ evenly distributed points in $\mathcal{U}$, and denote these points as $\mathbf{p}_i,i\in [N]=1,...,N$. For numerical application, we take $N$ to be a sufficiently large number and use the standard Monte-Carlo method \cite{robert1999monte} to approximate the evenly distributed points. 

In our implementation, we did not directly define $\epsilon$ due to the non-differentiability of the binarization required to determine if a $\mathbf{p}_i$ is within the $\epsilon$-tube. Instead, for each $\mathbf{p}_i$, we maximize the cosine similarity between $\mathbf{p}_i$ and its closest point on the trace. In this way, we relax the step function represented by \eqref{eq4} to its ``soft" version, leading to smooth gradient computation.

\subsection{Homogeneity}
\label{sec3.3}

While the enveloping loss effectively governs the overall distribution of representations on the hypersphere, it alone may not be entirely adequate, presenting two unresolved issues. 1) The first is distribution along the trace. The enveloping loss predominantly controls the overall shape of representations on the hypersphere, yet it does not guarantee a uniform distribution along the trace. This poses a notable concern, as it may result in uneven representation density across different trace segments. 2) The second is trace smoothness. The enveloping loss could lead to a zigzag pattern of the representations, which should be avoided. Considering age estimation from facial images as an example, the progression of facial features over time is gradual. Consequently, in the corresponding latent space, we would anticipate a similar, smooth transition without abrupt changes, underlining the desirability of a smoother trace. Interestingly, these two issues can be aptly analogized to winding yarns around a ball as well. For the yarn on the ball to be smooth, it should be tightly stretched, rather than being disjointed or loosely arranged. We name the property of a trace to be smooth with representations evenly distributed along it as \emph{homogeneity}.

We encourage such homogeneity property, \emph{i.e.}, smoothness of the trace $\mathrm{Im}(l)$ and uniform distribution of representations along it, by penalizing the arc length. Formally, the homogeneity loss is defined as:
\begin{equation}
\mathcal{L}_{\text{homo}}=\int_{y_{\mathrm{min}}}^{y_{\mathrm{max}}} \left\|\frac{\mathrm{d}l(y)}{\mathrm{d}y}\right\|^2 \mathrm{d}y
\end{equation}
Given $K$ different $y$s which have been ordered, the discrete format for $\mathcal{L}_{\text{homo}}$ is defined as a summation of the squared differences between adjacent points:
\begin{equation}
    \mathcal{L}_{\text{homo}} = \sum_{k=1}^{K-1} \frac{\left\|l(y_{k+1}) - l(y_k)\right\|^2}{y_{k+1}-y_{k}}
\end{equation}

The use of only homogeneity loss might result in trivial solutions like representation convergence to a circle or point due to feature collapse (shown in Figure \ref{fig2}). The homogeneity loss should be treated as a regularization of the enveloping loss, promoting not only smoothness but also an even distribution of representations along the trace. To quantitatively define the relationship between trace arc length and these desired characteristics, we introduce Theorem \ref{thm1}. It demonstrates that with a given $\text{Im}(l)$ ($l_{\text{env}}$ is fixed as it does not depend on the parameterization of $l$), the homogeneity loss is minimized if and only if when representations are uniformly distributed \emph{along} the trace.

\begin{theorem}
\label{thm1}
Given an image of $l$, $\mathcal{L}_{\text{homo}}$ attains its minimum if and only if the representations are uniformly distributed along the trace, i.e., $\left\|\nabla_yl(y)\right\|=c$, where $c$ is a constant.
\end{theorem}

Refer to Supplementary Material \ref{proof} for the proof.

Therefore, we formulate our geometric constraints ($\mathcal{L}_\text{G}$) as a combination of enveloping and homogeneity:

\begin{equation}
    \mathcal{L}_\text{G} = \lambda_{e}\mathcal{L}_{\text{env}} + \lambda_{h}\mathcal{L}_{\text{homo}}
\end{equation}

where $\lambda_{e}$ and $\lambda_{h}$ are weights for the two geometric losses. In Section \ref{sec4.4}, we further explore the behavior of these two geometric constraints, uncovering new insights into imbalanced regression.

\begin{figure}[t]
    \centering
    \includegraphics[width=\linewidth]{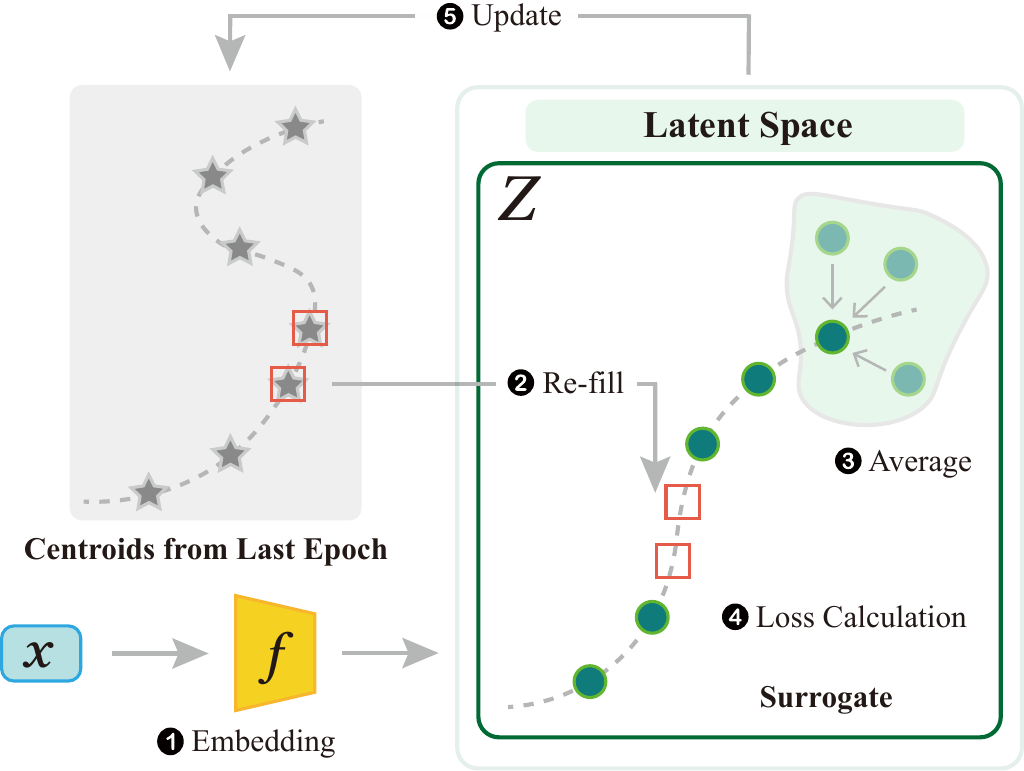}
    \caption{Overview of Surrogate-driven Representation Learning (SRL). \textbf{(1)} Every mini-batch is encoded to the latent space. Some bins may not be present in the current batch. To address this, \textbf{(2)} it takes centroids corresponding to the missing bins from the previous epoch. These stored centroids are used to ``re-fill" the missing bins in the current batch. \textbf{(3)} Average the representations for bins that appear multiple times, creating centroids for these bins. This surrogate, containing a representation for the full label range, allows for the effective application of geometric loss across all bins. \textbf{(4)} Loss calculation based on the surrogate. \textbf{(5)} Update the surrogate in memory to ensure enveloping and homogeneity. The training of the first epoch is driven by MSE loss only.}
    \label{fig4}
    \vspace{-0.4cm}
\end{figure}

\subsection{Surrogate-driven Representation Learning (SRL)}
\label{sec3.4}

Our geometric loss ($\mathcal{L}_\text{G}$) is calculated on a surrogate instead of a mini-batch (Figure~\ref{fig4}), as the representations from one mini-batch very likely fail to capture the global image of $l$, due to the randomness of batch sampling. 

For illustration purposes, here we assume the original dataset has already been binned, as is the case in most DIR datasets  \cite{yang2021delving,gong2022ranksim,keramati2023conr}. Let \(\mathcal{Z}=\{\mathbf{z}_1, \mathbf{z}_2,..., \mathbf{z}_M\}\) be a set of representations from a batch with batch-size \(M\), and let \(\mathcal{Y}=\{y_1, y_2,..., y_M\}\) (repetitions of the label values might exist) be the corresponding labels. Define the centroid \(\mathbf c_y \in \mathcal{C}\) for \(y\) ($\left\|\mathbf{c}_y\right\|=1$) as:
\begin{equation}
\mathbf c_y = \frac{1}{|\{\mathbf{z}_m \mid y_m = y\}|} \sum_{\{\mathbf{z}_m \mid y_m = y\}} \mathbf{z}_m
\end{equation}    

Suppose the whole label range is covered by a set of unique \(K\) bins \(\mathcal{Y}^* = \{y_1^*, y_2^*,..., y_K^*\}\). The centroids for these \(K\) bins from the last epoch are denoted as \(\mathcal{C'}=\{\mathbf{c'}_{y_1^*}, \mathbf{c'}_{y_2^*},..., \mathbf{c'}_{y_K^*}\}\). The surrogate \(\mathcal{S}\) is then generated by re-filling the missing centroids:
\begin{equation}
    \mathcal{S}=\mathcal{C}\cup \{\mathbf{c'}_{y_k^*} \mid y_k^* \in \mathcal{Y}^* \setminus \mathcal{Y} \}
\end{equation}

We use AdamW \cite{loshchilov2018decoupled} with momentum to update parameters $\theta$ in $f(\cdot)$, to ensure a smooth transition of the local shape in the batch-wise representations.

At the end of each epoch \( e \in (0,E] \) (excluding the first), we use the representations learned during that epoch to form a running surrogate \( \hat{\mathcal{S}}^e \).  \( \mathcal{S}^{e+1} \) is formulated from the current epoch's surrogate \( \mathcal{S}^{e} \)  and \( \hat{\mathcal{S}}^e \) with momentum. \( \mathcal{S}^{e+1} \) is employed for training in the subsequent epoch. It facilitates a gradual transition between epochs, preventing abrupt variations: $
    \mathcal{S}^{e+1} \leftarrow \alpha \cdot \mathcal{S}^{e} + (1-\alpha) \cdot \hat{\mathcal{S}}^{e}
$.

We aim for individual representations from the encoder to converge towards their respective centroids that share the same label, while simultaneously distancing them from centroids associated with different labels. To achieve this, we incorporate a contrastive loss between the individual representations and the centroids.
For each representation \(\mathbf{z}_m\) with $y_m=y$, the centroid $\mathbf c_y$ is considered as the positive, and the other centroids as negatives. The contrastive loss is defined as:
\begin{equation}
\mathcal{L}_{\text{con}} = -\sum_{m=1}^{M}{\log \frac{\exp(\text{sim}(\mathbf{z}_m, \mathbf c_{y}))}{\sum_{y^* \in \mathcal{Y}^*} \exp(\text{sim}(\mathbf{z}_{m}, \mathbf c_{y^*}))}}
\end{equation}
where $\text{sim}(\cdot)$ is the cosine similarity between two input.

The framework is trained end-to-end, the total loss used to update the parameters $\theta$ in $f(\cdot)$ is defined as:
\begin{equation}
    \mathcal{L}_{\theta}=\mathcal{L}_{\text{reg}}+\mathcal{L}_{\text{G}}+\mathcal{L}_{\text{con}}
\end{equation}

where $\mathcal{L}_{\text{reg}}$ is the mean squared error (MSE) loss.

\begin{table*}[htbp]
\centering
\caption{Results on UCI-DIR (MAE). We report the average MAE of three runs. The best results are in \textbf{bold}.}
\setlength{\tabcolsep}{3pt} 
\scriptsize 
\begin{tabular}{l|cccc|cccc|cccc|cccc}
\toprule
Datasets & \multicolumn{4}{c|}{Airfoil} & \multicolumn{4}{c|}{Abalone} & \multicolumn{4}{c|}{Real Estate} & \multicolumn{4}{c}{Concrete}\\ 
\midrule
Shot & All & Many & Med & Few & All & Many & Med & Few & All & Many & Med & Few & All & Many & Med & Few\\ 
\midrule
VANILLA          & 5.66 & 5.11 & 5.03 & 6.75 & 4.57 & \textbf{0.88} & 2.65 & 7.97 & 0.33 & 0.27 & 0.38 & 0.37 & 7.29 & 5.77 & 6.92 & 9.74\\
LDS + FDS    \citep{yang2021delving}    & 5.76 & \textbf{4.45} & 4.79 & 7.79 & 5.09 & 0.90 & 3.26 & 9.26 & 0.35 & 0.33 & 0.40 & 0.34 & 6.88 & 6.21 & 6.73 & 7.59\\
RankSim    \citep{gong2022ranksim}      & 5.23 & 5.05 & 4.91 & 5.72 & 4.33 & 0.98 & 2.59 & 7.42 & 0.37 & 0.34 & 0.38 & 0.40 & 6.71 & 6.00 & 5.57 & 9.46   \\
BalancedMSE   \citep{ren2022balanced}   & 5.69 & 4.51 & 5.04 & 7.28 & 5.37 & 2.14 & 2.66 & 9.37 & 0.34 & 0.31 & 0.40 & 0.33 & 7.03 & \textbf{4.67} & 6.37 & 9.72\\
Ordinal Entropy \citep{zhang2023improving} & 6.27 & 4.85 & 5.37 & 8.32 & 6.77 & 2.31 & 4.01 & 11.61 & 0.34 & 0.29 & 0.42 & 0.35 & 7.12 & 5.50 & 6.36 & 9.31\\  \midrule
SRL (ours)       & \textbf{5.10} & 4.83 & \textbf{4.75} & \textbf{5.69} & \textbf{4.16} & 0.89 & \textbf{2.42} & \textbf{7.19} & \textbf{0.28} & \textbf{0.26} & \textbf{0.30} & \textbf{0.29} & \textbf{5.94} & 5.32 & \textbf{5.80} & \textbf{6.60}\\
\toprule
\end{tabular}
\label{tab:table1}
\vspace{-0.4cm}
\end{table*}

\begin{table}[htbp]

\centering
\begin{threeparttable}
\caption{Results on AgeDB-DIR, the best are in \textbf{bold}.}
\scriptsize 
\setlength{\tabcolsep}{3pt} 
\renewcommand{\arraystretch}{1.1} 

\begin{tabular}{l|cccc|cccc}
\toprule
Metrics & \multicolumn{4}{c|}{MAE $\downarrow$} & \multicolumn{4}{c}{GM $\downarrow$} \\ \midrule
Shot & All & Many & Med & Few & All & Many & Med & Few \\ \midrule
VANILLA & 7.67 & 6.66 & 9.30 & 12.61 & 4.85 & 4.17 & 6.51 & 8.98 \\
LDS + FDS \citep{yang2021delving} & 7.55 & 7.03 & 8.46 & 10.52 & 4.86 & 4.57 & 5.38& 6.75 \\
RankSim \citep{gong2022ranksim} & 7.41 & \textbf{6.49} & 8.73 & 12.47 & 4.71 & 4.15 & 5.74 & 8.92 \\
BalancedMSE \citep{ren2022balanced}  & 7.98 & 7.58 & 8.65 & 9.93 & 5.01 & 4.83 & 5.46& 6.30\\
Ordinal Entropy  \citep{zhang2023improving} & 7.60& 6.69& 8.87& 12.68& 4.91& 4.28& 6.20& 9.29\\
ConR  \citep{keramati2023conr} & 7.41 & 6.51 & 8.81 & 12.04 & 4.70 & 4.13 & 5.91 & 8.59 \\ \midrule
SRL (ours) & \textbf{7.22} & 6.64 & \textbf{8.28} & \textbf{9.81} & \textbf{4.50} & \textbf{4.12} & \textbf{5.37} & \textbf{6.29}\\
\toprule
\end{tabular}
\vspace{-0.2cm}
\label{tab:table2}

\end{threeparttable}
\end{table}

\begin{table}[htbp]

\centering
\begin{threeparttable}
\caption{Results on IMDB-WIKI-DIR, the best are in \textbf{bold}.}
\scriptsize 
\setlength{\tabcolsep}{3pt} 
\renewcommand{\arraystretch}{1.1} 

\begin{tabular}{l|cccc|cccc}
\toprule
Metrics & \multicolumn{4}{c|}{MAE $\downarrow$} & \multicolumn{4}{c}{GM $\downarrow$} \\ \midrule
Shot & All & Many & Med & Few & All & Many & Med & Few \\ \midrule
VANILLA & 8.03& 7.16& 15.48& 26.11& 4.54& 4.14& 10.84& 18.64\\
LDS + FDS \citep{yang2021delving}  & 7.73 & 7.22& 12.98& 23.71& 4.40& 4.17& 7.87& 15.77\\
RankSim \citep{gong2022ranksim}  & 7.72& \textbf{6.92}& 14.52& 25.89& 4.29& 3.92& 9.72& 18.02\\
BalancedMSE \citep{ren2022balanced} & 8.43& 7.84& 13.35& 23.27& 4.93& 4.68& 7.90& 15.51\\
Ordinal Entropy \citep{zhang2023improving} & 8.01& 7.17& 15.15& 26.48& 4.47& 4.07& 10.56& 21.11\\
ConR \citep{keramati2023conr}  & 7.84& 7.15& 14.36& 25.15& 4.43& 4.05& 9.91& 18.55\\ \midrule
SRL (ours)& \textbf{7.69}& 7.08& \textbf{12.65}& \textbf{22.78}& \textbf{4.28}& \textbf{4.03}&\textbf{ 7.28}& \textbf{15.25}\\
\toprule
\end{tabular}
\vspace{-0.2cm}
\label{tab:table3}
\end{threeparttable}

\end{table}

\begin{table}[htbp]

\centering
\caption{Results on STS-B-DIR, the best are in \textbf{bold}.}
\scriptsize 
\setlength{\tabcolsep}{3pt} 
\renewcommand{\arraystretch}{1.1} 

\begin{tabular}{l|cccc|cccc}
\toprule
Metrics & \multicolumn{4}{c|}{MSE $\downarrow$} & \multicolumn{4}{c}{Pearson correlation $\uparrow$} \\ \midrule
Shot & All & Many & Med & Few & All & Many & Med & Few \\ \midrule
VANILLA & 0.993& 0.963& 1.000& 1.075& 0.742& 0.685& 0.693& 0.793\\
LDS + FDS \citep{yang2021delving}  & 0.900& 0.911& 0.881 & 0.905& 0.757& 0.698& 0.723& 0.806\\
RankSim  \citep{gong2022ranksim} & 0.889& 0.907& 0.874& 0.757& 0.763& 0.708& 0.692& 0.842\\
BalancedMSE  \citep{ren2022balanced} & 0.909& 0.894& 1.004& 0.809& 0.757& 0.703& 0.685& 0.831\\ 
Ordinal Entropy \citep{zhang2023improving} & 0.943 & 0.902 & 1.161 & 0.812 & 0.750 & 0.702 & 0.679 & 0.767  \\ \midrule
SRL (ours)& \textbf{0.877}& \textbf{0.886}& \textbf{0.873}& \textbf{0.745}& \textbf{0.765}& \textbf{0.708}& \textbf{0.749}& \textbf{0.844}\\
\toprule
\end{tabular}
\vspace{-0.2cm}
\label{tab:table4}
\end{table}

\section{Experiments}

We perform extensive experiments to validate and analyze the effectiveness of SRL for deep imbalanced regression. Our regression tasks span age estimation from facial images, tabular regression, and text similarity score regression, as well as our newly established task: Imbalanced Operator Learning (IOL). This section begins by detailing the experiment setup (Section \ref{sec4.1}) followed by the main results (Section \ref{sec4.2}). The results of IOL are shown in Section \ref{sec4.3} followed by the comparison with classification-based methods and hyperparameters analysis (Section \ref{sec4.4}).

\subsection{Experiment Setup}
\label{sec4.1}

\textbf{Datasets.} We employ three real-world regression datasets developed by \citet{yang2021delving}, and our curated UCI-DIR from UCI Machine Learning Repository \cite{asuncion2007uci}, to assess the effectiveness of SRL in deep imbalanced regression. Refer to Supplementary Material \ref{sec7} for more dataset details.

\begin{itemize}
    \item \textbf{\emph{AgeDB-DIR}} \cite{yang2021delving}: It serves as a benchmark for estimating age from facial images, which is derived from the AgeDB dataset \cite{moschoglou2017agedb}. It contains 12,208 images for training, 2,140 images for validation, and 2,140 images for testing. 
    \item \textbf{\emph{IMDB-WIKI-DIR}} \cite{yang2021delving}: It is a facial image dataset for age estimation derived from the IMDB-WIKI dataset \cite{rothe2018deep}, which consists of face images with the corresponding age. It has 191,509 images for training, 11,022 images for validation, and 11,022 for testing.
    \item \textbf{\emph{STS-B-DIR}} \cite{yang2021delving}: It is a natural language dataset formulated from STS-B dataset \cite{cer2017semeval,wang2018glue}, consisting of 5,249 training sentence pairs, 1,000 validation pairs, and 1,000 testing pairs. Each sentence pair is labeled with the continuous similarity score.
    \item \textbf{\emph{UCI-DIR}}: To evaluate the performance of SRL on tabular data, we curated UCI Machine Learning Repository \cite{asuncion2007uci} to formulate UCI-DIR that includes four regression tasks (Airfoil Self-Noise, Abalone, Concrete Compressive Strength, Real estate valuation). Following the DIR setting \cite{yang2021delving}, we make each regression task consist of an imbalanced training set and a balanced validation and test set.
\end{itemize}

\noindent\textbf{Metrics.} In line with the established settings in DIR \citep{yang2021delving}, subsets in an imbalanced training set are categorized based on the number of available training samples: many-shot region (bins with $>$ 100 training samples), medium-shot region (bins with 20 to 100 training samples), and few-shot region (bins with $<$ 20 training samples), for the three real-world datasets. For AgeDB-DIR and IMDB-WIKI-DIR, each bin represents 1 year. In the case of STS-B-DIR, bins are segmented by 0.1. For UCI-DIR, the bins are segmented by 0.1 to 1 depending on the range of regression targets. Our evaluation metrics include mean absolute error (MAE, the lower the better) and geometric mean (GM, the lower the better) for AgeDB-DIR, IMDB-WIKI-DIR and UCI-DIR. For STS-B-DIR, we use mean squared error (MSE, the lower the better) and Pearson correlation (the higher the better). 

\noindent\textbf{Implementation Details.} For age estimation (AgeDB-DIR and IMDB-WIKI-DIR), we follow the settings from \citet{yang2021delving}, which uses ResNet-50 \cite{he2016deep} as a backbone network. For text similarity regression (STS-B-DIR), we follow the setting from \citet{cer2017semeval,yang2021delving} that uses BiLSTM + GloVe word embeddings. For tabular regression (UCI-DIR), we use an MLP with three hidden layers (d-20-30-10-1) following the setting from \citet{cheng2023weakly}.  For all baseline methods, results were produced following provided training recipes through publicly available codebase. All experimental results, including ours and baseline methods, were obtained from a server with 8 RTX 3090 GPUs.

\noindent\textbf{Baselines.} We consider both DIR methods \citep{yang2021delving,gong2022ranksim,keramati2023conr} and recent techniques proposed for general regression \citep{ren2022balanced,zhang2022improving,zhang2023improving}, in addition to VANILLA regression (MSE loss). We compare the performance of SRL with all baselines on the above four datasets. Furthermore, as SRL is orthogonal to previous DIR methods, we examine the improvement of them by adding our geometric losses.

\begin{table}[htbp]
\centering
\caption{Combine SRL with existing DIR methods (MAE)}
\scriptsize 
\setlength{\tabcolsep}{2pt} 
\renewcommand{\arraystretch}{1.1} 
\begin{tabular}{l|cccc|cccc}
\toprule
Datasets & \multicolumn{4}{c|}{AgeDB (MAE)} & \multicolumn{4}{c}{IMDB-WIKI (MAE)} \\ \midrule
Shot & All & Many & Med & Few & All & Many & Med & Few \\ \midrule
SRL+LDS+FDS \citep{yang2021delving} & 7.32 & 6.81 & 8.14 & 9.81 & 7.61 & 7.03 & 12.28 & 21.77 \\
GAINS v.s. LDS+FDS (\%) & \textcolor{darkergreen}{3.05} & \textcolor{darkergreen}{3.23} & \textcolor{darkergreen}{3.89} & \textcolor{darkergreen}{6.75} & \textcolor{darkergreen}{1.66} & \textcolor{darkergreen}{2.64} & \textcolor{darkergreen}{5.40} & \textcolor{darkergreen}{8.19} \\ \midrule
SRL+RankSim \citep{gong2022ranksim} & 7.29 & 6.57 & 8.58 & 10.48 & 7.67 & 7.08 & 12.40 & 22.85 \\
GAINS v.s. RankSim (\%)& \textcolor{darkergreen}{1.62} & \textcolor{blue}{-1.23} & \textcolor{darkergreen}{1.72} & \textcolor{darkergreen}{16.96}& \textcolor{darkergreen}{0.65} & \textcolor{blue}{-1.15} & \textcolor{darkergreen}{14.61} & \textcolor{darkergreen}{11.75} \\ \midrule
SRL+BalancedMSE \citep{ren2022balanced} & 7.24 & 6.77 & 7.86 & 9.85 & 7.74 & 7.13 & 12.77 & 22.04 \\
GAINS v.s. BalancedMSE (\%)& \textcolor{darkergreen}{9.27} & \textcolor{darkergreen}{10.69} & \textcolor{darkergreen}{9.14} & \textcolor{darkergreen}{0.89} & \textcolor{darkergreen}{8.19} & \textcolor{darkergreen}{9.06} & \textcolor{darkergreen}{4.35} & \textcolor{darkergreen}{5.29} \\ \midrule
SRL+ConR \citep{keramati2023conr} & 7.40 & 6.87 & 8.08 & 10.50 & 7.56 & 7.01 & 12.03 & 21.71 \\
GAINS v.s. ConR (\%)& \textcolor{darkergreen}{0.14} & \textcolor{blue}{-5.53} & \textcolor{darkergreen}{8.39} & \textcolor{darkergreen}{13.80}& \textcolor{darkergreen}{3.68} & \textcolor{darkergreen}{1.96} & \textcolor{darkergreen}{16.23} & \textcolor{darkergreen}{13.68} \\
\bottomrule
\end{tabular}
\label{tab:table5}
\vspace{-0.4cm}
\end{table}

\begin{figure*}
    \centering
    \includegraphics[width=0.9\linewidth]{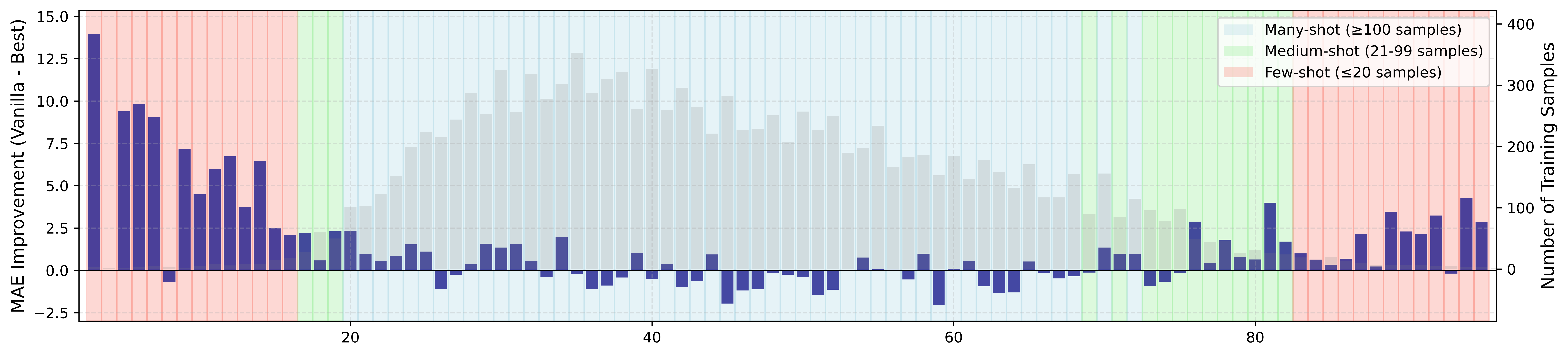}
    \caption{SRL performance gain compared to VANILLA across age ranges on AgeDB-DIR. The gray histogram in the background shows the distribution of samples across age groups. SRL substantially improves the performance on the medium-shot and few-shot regions (age $<$ 20 and $>$ 70).}
    \vspace{-0.4cm}
    \label{fig:fig9}
\end{figure*}

\subsection{Main Results}
\label{sec4.2}

To show the effectiveness of SRL on DIR, we first benchmark SRL and baselines for tabular regression on our curated UCI-DIR with four different regression tasks (Table \ref{tab:table1}). Moreover, we evaluate our method on established DIR benchmarks \citep{yang2021delving} including age estimation on AgeDB-DIR and IMDB-WIKI-DIR (Table \ref{tab:table2} \& \ref{tab:table3}, and Figure \ref{fig:fig9}), and text similarity regression on STS-B-DIR (Table \ref{tab:table4}). We evaluate the combination of SRL and previous DIR methods on AgeDB-DIR and IMDB-WIKI-DIR (Table \ref{tab:table5}). Notably, Table \ref{tab:table1} and \ref{tab:table4} omit results from ConR \citep{keramati2023conr}, as it depends on data augmentation, a technique not fully established in the domain of tabular data and natural language.

\noindent\textbf{Combine SRL with existing methods.} Our SRL approach enhances imbalanced regression by imposing geometric constraints on feature distributions, a strategy that is orthogonal to existing methods. To illustrate this, we leverage SRL as a regularizing term in conjunction with other methods. The results of this experiment are presented in Table \ref{tab:table5}. It shows that when SRL is integrated with existing regression methods, there is improvement in performance across different regions for both datasets. This demonstrates the effectiveness and compatibility of SRL as a complementary tool in the realm of regression analysis.

\subsection{Imbalanced Operator Learning (IOL)}
\label{sec4.3}

\begin{figure}[t]
    \centering
    \includegraphics[width= 0.6\columnwidth]{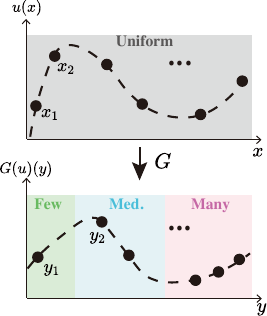}
    \vspace{-0.3cm}
    \caption{Imbalanced Operator Learning.}

    \label{fig5}
\end{figure}

We introduce a novel task for DIR called Imbalanced Operator Learning (IOL). Traditional operator learning aims to train a neural network to model the mapping between function spaces \cite{kovachki2023neural,lu2021learning}. However, unlike the standard approach of uniformly sampling output locations, in IOL, we intentionally adjust the sampling density within the output function's domain to create regions with few, medium, and many regions (Figure \ref{fig5}).

For the linear operator, the model is trained to estimate the integral operator denoted as $G$:
\begin{equation}
    G: u(x) \mapsto s(x)=\int_{0}^{x}u(\tau) d\tau, x\in[0,1]
\end{equation}
\label{eq1}

where \(u\) denotes the input function which is sampled from a Gaussian random field (GRF), and \(s\) is the target function. 

For the nonlinear operator, the model is trained to learn a particular stochastic partial differential equation (PDE):
\begin{equation}
\operatorname{div}\left(e^{b(x ; \omega)} \nabla u(x ; \omega)\right) = f(x), x\in[0,1]
\end{equation}
\label{eq2}
where $e^{b(x ; \omega)}$ is the diffusion efficient and $u(x ; \omega)$ is the target function. 

Denote the domain of output function as $y$. For both linear and non-linear operator learning,  we changed the original uniform sampling of \(y\) to three curated regions: few/medium/many. Afterward, we manually created an imbalanced training set of 10k samples and a balanced testing test of 100k samples, namely OL-DIR. 

In Figure \ref{fig5}, we have a schematic overview of Imbalanced Operator Learning (IOL). The network is trained to model an integral operator $G$. The data provided to the model is $([u,y],G(u)(y))$. The input consists of function $u$ and sampled $y$s from the domain of $G(u)$. The target is $G(u)(y)$. We manipulate the distribution density of $y$ across its range to formulate few/med./many regions. Here the imbalance comes from the unequal exposure of integral interval to the model training. Refer to Supplementary Material \ref{sec.c.2} for more details.

Shown in Table \ref{tab:table6}, SRL consistently outperforms VANILLA and the state-of-the-art operator learning for the whole label range including all, many-shot, medium-shot, and few-shot regions. The results position SRL as the superior approach for IOL in terms of accuracy and generalizability.

\begin{table}[htbp]
\centering
\caption{Results on OL-DIR. We report the average MAE of ten runs. The best results are \textbf{bold}.}
\scriptsize 
\setlength{\tabcolsep}{3pt} 
\renewcommand{\arraystretch}{1.1} 
\begin{tabular}{l|cccc|cccc}
\toprule
Operation & \multicolumn{4}{c|}{ MAE($10^{-3}$)  $\downarrow$} & \multicolumn{4}{c}{ MSE ($10^{-4}$) $\downarrow$} \\ \midrule
Shot & All & Many & Med & Few & All & Many & Med & Few \\ \midrule
\textit{\textbf{Linear}} \\ \midrule
VANILLA \citep{lu2021learning}& 15.64 &  11.86 & 15.45 & 27.00 & 5.40 & 2.81 & 4.40 & 14.20 \\
Ordinal Entropy \citep{zhang2023improving}& 10.07  & 9.26 & 9.85 & 13.01 & 2.00 & 1.53 & 1.89 & 3.42 \\ 
SRL (ours)  & \textbf{9.18} & \textbf{8.32} & \textbf{9.47} & \textbf{9.33}  & \textbf{1.98} & \textbf{0.98} & \textbf{1.72} & \textbf{2.67} \\ \midrule
\textit{\textbf{Nonlinear}} \\ \midrule
VANILLA \citep{lu2021learning} & 11.64 & 9.89 & 11.02 & 19.77 & 9.20 & \textbf{4.33} & 7.53 & 24.70 \\
Ordinal Entropy \citep{zhang2023improving} & 12.91 & 9.93 & 13.07 & 21.02 & 13.80 & 8.82 & 11.84 & 30.12 \\
SRL (ours)  & \textbf{11.25} & \textbf{9.48} & \textbf{9.22} &  \textbf{17.00} &  \textbf{8.60} & 7.42 & \textbf{6.41} & \textbf{14.12} \\
\toprule

\end{tabular}
\par{Full results with standard deviation are reported in Supplementary Material \ref{ol_result}.}

\label{tab:table6}
\vspace{-0.5cm}
\end{table}

\subsection{Further Analysis}
\label{sec4.4}

\textbf{Quantification of geometric impact.} We further quantified the impact of geometric constraints by comparing percentages of uniformly sampled points within few-shot regions (a measure of proportion). The results show our method significantly increases few-shot proportion (AgeDB-DIR: $1.98\%\rightarrow15.80\%$, upper bound: 23\%; STS-B-DIR: $4.52\%\rightarrow22.39\%$, upper bound: 38\%), leading to improved performance (Table \ref{proportion}).

\begin{table}[htbp]
    \centering
    \caption{Impact of geometric constraints on few-shot proportion.}
    \resizebox{\columnwidth}{!}{
    \footnotesize
    \begin{tabular}{c@{\hspace{5em}}c@{\hspace{5em}}c@{\hspace{5em}}c}
        \toprule
        & \multicolumn{2}{c}{\hspace{-4em}Few-shot} & Overall \\
        \cmidrule(l{-0.3em}r{5em}){2-3} \cmidrule(l{-0.3em}r){4-4}
        & Proportion & MAE & MAE \\
        \midrule
        \multicolumn{4}{l}{\textbf{AgeDB-DIR (\emph{1.10\% samples, 23\% label range}):}} \\
        \cmidrule{1-4}
        VANILLA & 1.98\% & 12.61 & 7.67 \\
        LDS + FDS \citep{yang2021delving} & 4.95\% & 10.52 & 7.55 \\
        Ours & \textbf{15.80\%} & \textbf{9.81} & \textbf{7.22} \\
        \cmidrule{1-4}
        \multicolumn{4}{l}{\textbf{STS-B-DIR (\emph{3.49\% samples, 38\% label range}):}} \\
        \cmidrule{1-4}
        VANILLA & 4.52\% & 1.075 & 0.993  \\
        LDS + FDS \citep{yang2021delving} & 8.13\% & 0.905 & 0.900  \\
        Ours & \textbf{22.39\%} & \textbf{0.877} & \textbf{0.745} \\
        \bottomrule
    \end{tabular}}
    \label{proportion}
\end{table}

\begin{figure}[htbp]
    \centering
    \includegraphics[width=1.0\columnwidth]{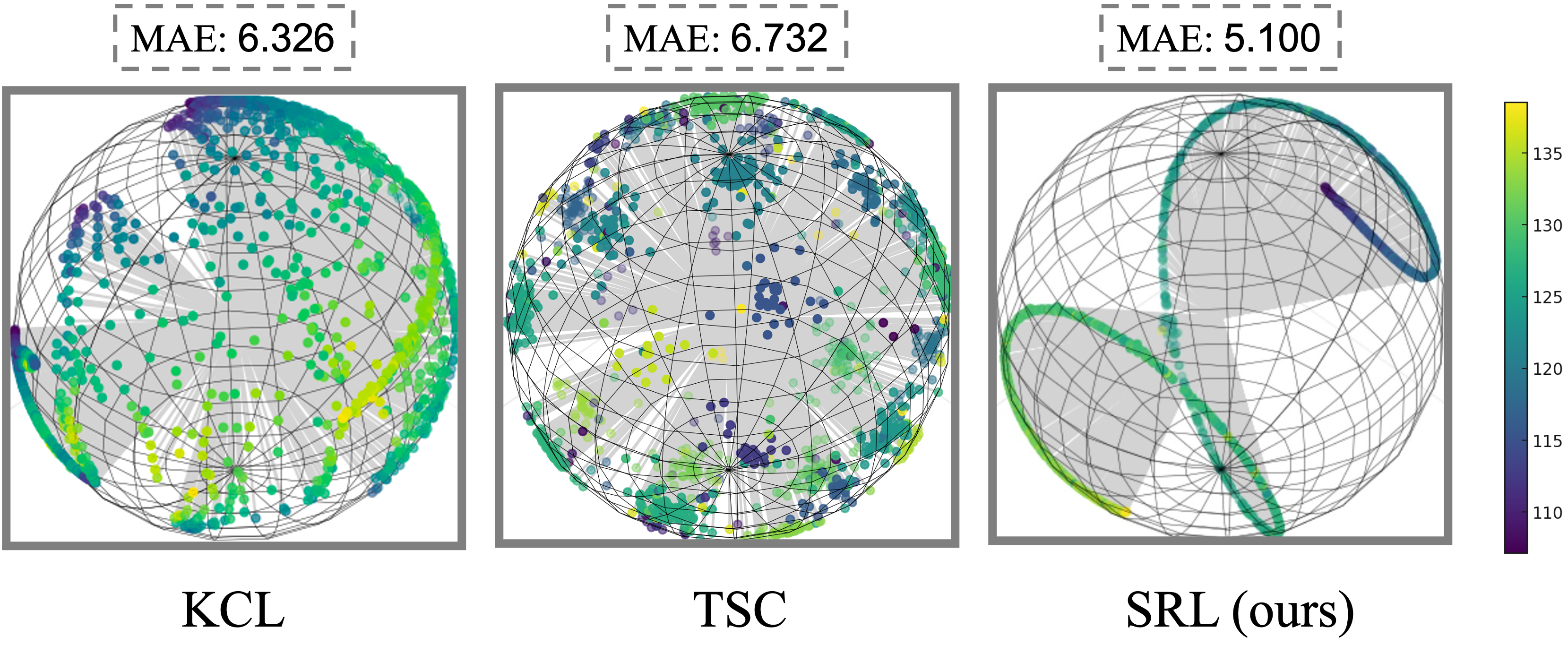}
    \vspace{-0.8cm}
    \caption{Comparison of the feature distributions among KCL \citep{kang2020exploring}, TSC \citep{li2022targeted} and SRL(ours) on UCI-Airfoil. All methods aim to promote uniformity in feature distribution while KCL and TSC are originally proposed for imbalanced classification.}
    \label{fig6}
\end{figure}

\textbf{Compare with methods for long-tailed classification:} In Figure \ref{fig6}, we compare the feature distribution of our method with KCL \citep{kang2020exploring} and  TSC \citep{li2022targeted}. This comparison reveals that classification-based approaches like KCL and TSC tend to distribute feature clusters on the hypersphere by positioning the target centroids at maximal distances from one another. However, this strategy adversely affects the ordinality and continuity which are essential for regression tasks. As a result, such methods often lead to suboptimal performance for imbalanced regression, even worse than any of the regression baselines shown in Figure \ref{fig1}.

\begin{figure}[htbp]
    \centering
    \vspace{-0.5cm}
    \includegraphics[width=0.8\columnwidth]{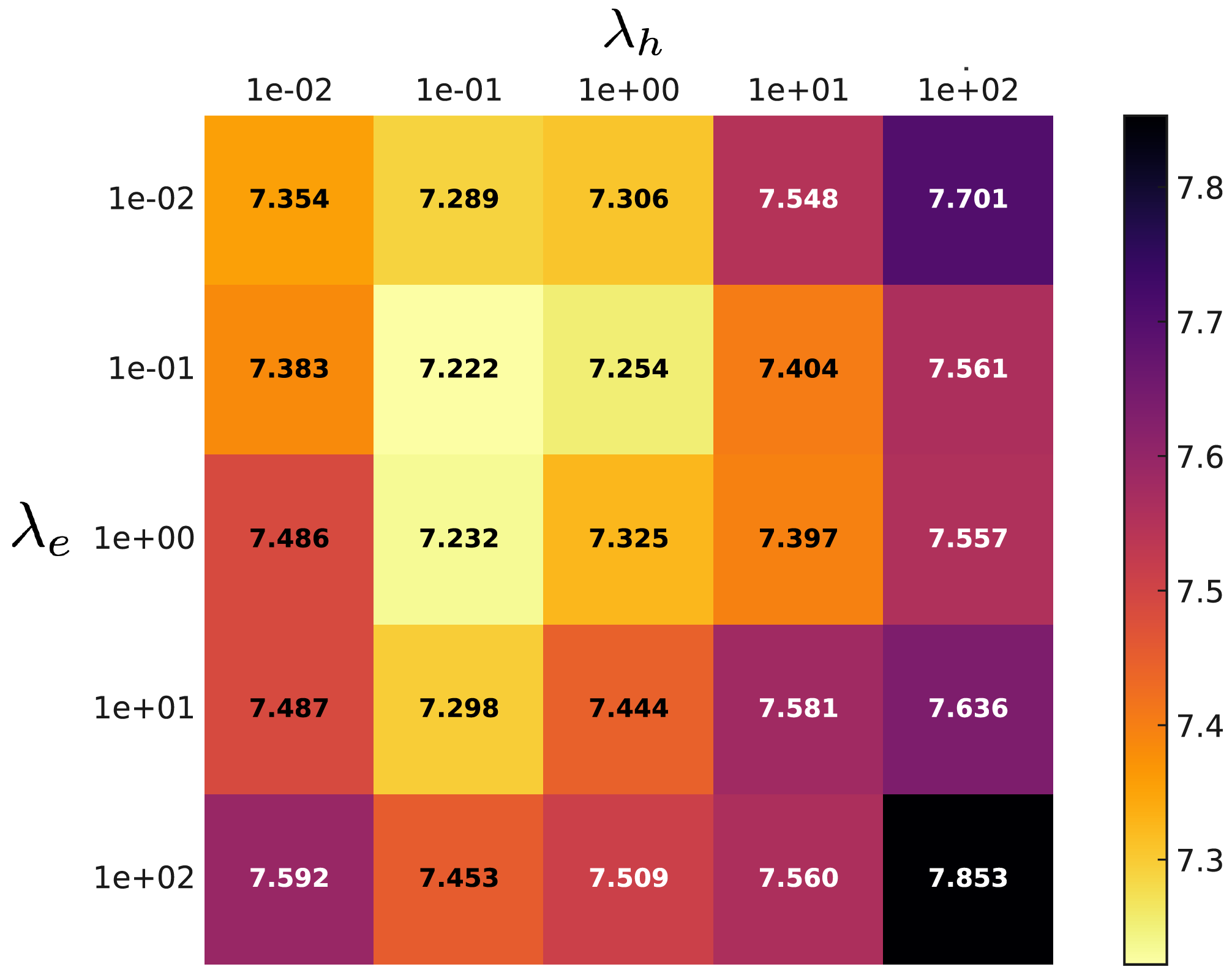}
    \captionof{figure}{Confusion matrix of MAE on AgeDB-DIR from different values of $\lambda_h$ and $\lambda_e$.}
    \label{fig7}
\end{figure}

\noindent\textbf{Balancing of enveloping and homogeneity:} Our proposed SRL advocates for two pivotal geometric constraints in feature distribution: enveloping and homogeneity, to effectively address imbalanced regression. These two losses are modulated by their respective coefficients, $\lambda_e$ for the enveloping loss and $\lambda_h$ for the homogeneity loss. Figure \ref{fig7} illustrates that the omission of either constraint detrimentally impacts the performance, highlighting the importance of both of them, and it demonstrates that the best performance, as measured by Mean Absolute Error (MAE) on the AgeDB-DIR dataset, is achieved when both coefficients $\lambda_e$ and $\lambda_h$ are set to $1e^{-1}$.

\noindent\textbf{Ablation studies on choices of $N$:} Table \ref{tab:table8} (in Supplementary Material) shows that achieving optimal performance on the AgeDB-DIR and IMDB-WIKI-DIR datasets requires a sufficiently large $N$, as a smaller $N$ may lead to imprecise calculation of the enveloping loss.

\noindent\textbf{Ablation studies on proposed loss component:} Table \ref{tab:table7} (in Supplementary Material) demonstrates that incorporating homogeneity, enveloping, and contrastive loss term yields superior model performance compared to using each individually.

\noindent\textbf{Computational cost:} As shown in Table \ref{tab:training_time} (in Supplementary Material), the computational overhead introduced by the Surrogate-driven Representation Learning (SRL) framework is comparable to that of other imbalanced regression methods.

\noindent\textbf{Impact of bin numbers:} As shown in Table \ref{tab:bin_ablation} (in Supplementary Material), while increasing the number of bins generally leads to better model performance, the improvements become marginal beyond certain thresholds.

\noindent\textbf{Limitations:} Section \ref{F} (in Supplementary Material) examines the limitations of SRL, including its inability to handle higher-dimensional labels.

\vspace{-0.3cm}
\section{Conclusion}
As the first work of exploring uniformity in deep imbalanced regression, we introduce two novel loss functions - enveloping and homogeneity loss - to encourage the uniform feature distribution of an ordered and continuous trajectory. The two loss functions serve as geometric constraints which are integrated into the data representations through a Surrogate-driven Representation Learning (SRL) framework. Furthermore, we set a new benchmark in imbalanced regression: Imbalanced Operator Learning (IOL). Extensive experiments on real-world regression and operator learning demonstrate the effectiveness of our geometrically informed approach. We emphasize the significance of uniform data representation and its impact on learning performance in imbalanced regression scenarios, advocating for a more balanced and comprehensive utilization of feature spaces in regression models.

{
    \small
    \bibliographystyle{ieeenat_fullname}
    \bibliography{main}
}

\input{sec/X_suppl}

\end{document}

%% file: sec/X_suppl.tex
\clearpage
\setcounter{page}{1}
\maketitlesupplementary


\section{Proof of Theorem 1.}
\label{proof}

\begin{proof}
    Define $y \in [0,1]$.
A reparametrization of the path $l(y)$ is defined by a bijective strictly increasing function $r(y):[0,1]\rightarrow [0,1]$, denoted as $\Tilde{l}(y):=(l \circ r)(y)$. Due to the fact that $\mathrm{Im}(l) = \mathrm{Im}(\Tilde{l})$,
\begin{equation}
    T(l,\epsilon) = T(\Tilde{l},\epsilon') \Rightarrow \mathcal{L}_{\text{env}}(l,\epsilon) = \mathcal{L}_{\text{env}}(\Tilde{l},\epsilon')
\end{equation}
Denote $r'$ as the derivative of $r$. Further we have
\begin{equation}
   \begin{aligned}
\mathcal{L}_{\text{homo}}(\Tilde{l}) &= \int_0^1 |\nabla_y l(r(y))|^2 dy \\
&= \int_0^1|\nabla_r l(r)|^2|_{r=r(y)} \cdot |r'(y)|^2 dy \\
&= \int_0^1|\nabla_r l(r)|^2|_{r=r(y)} \cdot r'(y)^2 dy \\
&= \int_0^1|\nabla_r l(r)|^2 \cdot r'(y) dr \\
&= \int_0^1|\nabla_r l(r)|^2 \cdot s(r) dr,
\end{aligned} 
\end{equation}

where $s = r'\circ r^{-1}$. This separates the dependence of $\mathcal{L}_{\text{homo}}$ on the reparametrization to a single weight function $s:[0,1]\rightarrow \mathbb{R}_+$. 

Then we have
\begin{equation}
\mathcal{L}_{\text{homo}}(\Tilde{l})-\mathcal{L}_{\text{homo}}(l) = \int_0^1|\nabla_y l(y))|^2(s(y)-1)dy.
\end{equation}

Now if the original curve is moving at constant speed, \emph{i.e.}, $|\nabla_yl(y)| = c$, where $c$ is a positive constant. In other words, the data is uniformly distributed. Then
\begin{align*}
\mathcal{L}_{\text{homo}}(\Tilde{l}) - \mathcal{L}_{\text{homo}}(l) &= c^2 \int_0^1 (s(y) - 1) \, dy \\
&= c^2 \left( \int_0^1 s(y) \, dy - 1 \right),
\end{align*}

which means in this case the loss will increase if $\int_0^1s(y)dy>1$ and decrease otherwise. Since $r$ is a bijection, we have
\begin{align*}
\int_0^1 s(r)\,dr &= \int_0^1 s(r(y))r'(y)\,dy \\ &=\int_0^1 r'(y)^2\,dy
\end{align*}

Since $(r'(t)-r'(y))^2\geq 0$, $t,y\in [0,1]$, we have
\begin{align*}
0 &\leq \int_{0}^1\int_{0}^1 (r'(y) - r'(t))^2 \, dt \, dy \\
&= 2\int_0^1\int_0^1 r'(y)^2 \, dy \, dt - 2\left(\int_0^1 r'(y) \, dy\right)^2 \\
&= 2\int_0^1 r'(y)^2 \, dy - 2 \\
&\Rightarrow \int_0^1 r'(y)^2 \, dy \geq 1,
\end{align*}

where the inequality holds when $r'(y)$ is a constant, since $r$ is bijective, $r$ should be the function:  $r(y)=y$. This means $l(y) = \Tilde{l}(y),\forall y$. Therefore, we have $\int_0^1r'(y)^2dy> 1$, for $\Tilde{l}\neq l$, which means, the loss attains its minimum if and only if the data is uniformly distributed.
\end{proof}

\section{Datasets}
\label{sec7}

\subsection{UCI-DIR}
\label{uci_dir}
\begin{figure*}[htbp]
    \centering
    \includegraphics[width=1.6\columnwidth]{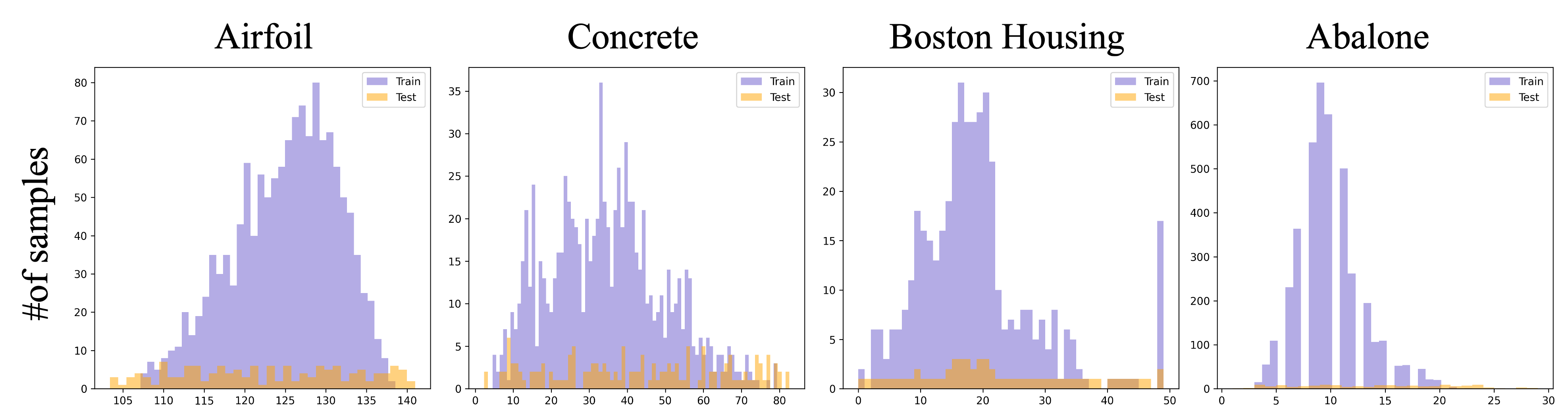}
    \caption{Overview of training and test set label distribution for UCI-DIR datasets. }
    \label{fig:uci_dist}
\end{figure*}

We curated UCI-DIR to evaluate the performance of imbalanced regression methods on tabular datasets. Here, we consider four regression tasks from UCI machine learning repository \citep{asuncion2007uci} (Airfoil, Concrete, Real Estate and Abaleone). Their input dimensions range from 5 to 8. Following the original DIR setting \citep{yang2021delving}, we curated a balanced test set with balanced distribution across the label range and leave the training set naturally imbalanced (Figure \ref{fig:uci_dist}). We partitioned the label range into three regions based on the occurrence. The threshold for [few-shot/med-shot, med-shot/many-shot] are [10, 40], [5, 15], [3, 10] and [100, 400] for Airfoil, Concrete, Real Estate and Abalone respectively.

\begin{table*}[htbp]
    \centering
    \scriptsize
    \caption{Overview of the six curated datasets used in our experiments}
    \begin{tabular}{c|cccccccc}
    \toprule Dataset & Target type & Target range & Bin size & \# Training set & \# Val. set & \# Test set \\
    \midrule IMDB-WIKI & Age & \(0 \sim 186\textit{*}\) & 1  & 191,509 & 11,022 & 11,022 \\
     AgeDB-DIR & Age & \(0 \sim 101\) & 1 & 12,208 & 2,140 & 2,140 \\
     STS-B-DIR & Text similarity score & \(0 \sim 5\) & 0.1 & 5,249 & 1,000 & 1,000 \\
    \toprule
    \end{tabular}
    \label{tab:real_dataset}
    \parbox[t]{0.6\textwidth}{\textit{*Note: wrong labels in the original dataset.}}
\end{table*}

\subsection{OL-DIR}
\label{sec.c.2}

We follow \citet{lu2021learning} for the basic setting of operator learning. However, we change the original uniform sampling of locations in the domain of the output function to three regions: few, medium, and many regions.

For the linear operator defined in Equation (12), the input function $u$ is generated from a Gaussian Random Field (GRF):

\begin{equation}
    u \sim \mathcal{G}(0, k(x_1, x_2))
\end{equation}

\begin{equation}
        k(x_1, x_2) = \exp\left(-\frac{\|x_1 - x_2\|^2}{2l^2}\right)
\end{equation}

where the length-scale parameter $l$ is set to be 0.2. For $x$, we fix 100 locations to represent the input function $u$. The locations in the output function $y$s are manually sampled from the domain of $G(u)$, such that few-shot region: $y\in [0.0,0.2]\cup [0.8,1.0]$; medium-shot region: $y\in [0.2,0.4]\cup [0.6,0.8]$; many-shot region: $y\in [0.4,0.6]$.

We manually create an imbalanced training set with many/medium/few-shot regions of 10k samples and a balanced testing test of 100k samples.

For the nonlinear operator defined in Equation (13), the input function is defined as:

\begin{equation}
    b(x; \omega) \sim \mathcal{GP}(b_0(x), \text{cov}(x_1, x_2))
\end{equation}

\begin{equation}
    b_0(x) = 0
\end{equation}

\begin{equation}
    \text{cov}(x_1, x_2) = \sigma^2 \exp\left(-\frac{\|x_1 - x_2\|^2}{2l^2}\right)
\end{equation}

 where $\omega$ is sampled from a random space with Dirichlet boundary conditions $u(0)=u(1)=0, \ f(x)=10$. $\mathcal{GP}$ is a Gaussian random process. The target locations are sampled in the same way as the linear task.

 The number and split of the nonlinear operator dataset are the same as those of the linear one.

\subsection{AgeDB-DIR, IMDB-DIR and STS-B-DIR}

For the real-world datasets (AgeDB-DIR, IMDB-WIKI-DIR and STS-B-DIR), We follow the original train/val./test split from \citep{yang2021delving}.

\label{B1}

\subsection{Ethic Statements}
\label{B2}
All datasets used in our experiments are publicly available and do not contain private information. 
All datasets (AgeDB, IMDB-WIKI, STS-B, and UCI) are accrued without any engagement or interference involving human participants and are devoid of any confidential information.

\section{Experiment Detail}
\label{d}

\subsection{Implementation Detail (Table \ref{tab:hyper_param}).}

\begin{table*}[htbp]
    \centering
    \scriptsize
    \caption{Hyper-parameters used in SRL}
    \begin{tabular}{c|ccccc}
    \toprule Dataset & {IMDB-WIKI} & {AgeDB-DIR} & {STS-B-DIR} & {UCI-DIR} & {OL-DIR}\\ \midrule
     {Temperature ($\tau$)} & 0.1 & 0.1 & 0.1  & 0.1 & 0.1 \\
     {Momentum ($\alpha$)} & 0.9 & 0.9 & 0.9 & 0.9 & 9.9 \\
     {N} & 2000 & 2000 & 1000 & 1000 & 1000 \\
     {$\lambda_{e}$} & 1e-1 & 1e-1 & 1e-2  & 1e-2 & 1e-1 \\
     {$\lambda_{h}$} & 1e-1 & 1e-1 & 1e-4  & 1e-2 & 1e-1 \\
     {Backbone Network $(f(\cdot))$} & ResNet-50 & ResNet-50 & BiLSTM & 3layer MLP & 3layer MLP \\
     {Feature Dim} & 128 & 128 &  128 & 128  & 128 \\
     {Learning Rate} & 2.5e-4 & 2.5e-4 & 2.5e-4  & 1e-3  & 1e-3 \\
     {Batch Size} & 256 & 64 & 16  & 256 & 1000 \\
    \toprule 
    \end{tabular}
    \label{tab:hyper_param}
\end{table*}

\subsection{Choices of $N$.}
\label{N}

We investigate how varying $N$ (the number of uniformly distributed points on a hypersphere used to calculate enveloping loss) impacts the performance of our approach on the AgeDB-DIR and IMDB-WIKI-DIR datasets (Table \ref{tab:table8}). To achieve optimal performance, it is crucial to choose a sufficiently large $N$. A smaller $N$  might fail to cover the entire hypersphere adequately, resulting in an imprecise calculation of enveloping loss.

\begin{table}[htbp]
\centering
\scriptsize
\caption{Vary the number of $N$}
\setlength{\tabcolsep}{3pt} 
\renewcommand{\arraystretch}{1.1} 
\begin{tabular}{l|ccccccc}
\toprule
$N$ & 100 & 200 & 500 & 1000 & \textbf{2000} & 4000 & 10000 \\ \midrule
AgeDB & 7.78 & 7.55 & 7.37 & 7.31 & 7.22 & 7.22 & 7.22 \\
IMDB-WIKI & 7.85 & 7.78 & 7.72 & 7.69 & 7.69 & 7.69 & 7.72\\
\toprule
\end{tabular}

\label{tab:table8}

\end{table}

\subsection{Ablation on proposed components.} The Table \ref{tab:table7} presents the results of an ablation study examining the impact of different loss functions on the model performance. As we mentioned before, the use of only homogeneity loss ($\mathcal{L}_{\text{homo}}$) could lead to trivial solutions due to feature collapse. Additionally, using only the enveloping loss ($\mathcal{L}_{\text{env}}$) causes the features to spread out along the trajectory, resulting in suboptimal performance. Through the contrastive loss ($\mathcal{L}_{\text{con}}$), individual representations could converge towards their corresponding locations on the surrogate. It is evident from the Table \ref{tab:table7} that the model incorporating all loss functions outperforms the other configuration. 

\begin{table}[htbp]
\centering
\scriptsize

\caption{Ablation Studies, best results are bold}
\setlength{\tabcolsep}{3pt} 
\renewcommand{\arraystretch}{1.1} 

\begin{tabular}{cc|c|cccc|cccc}
\toprule
$\mathcal{L}_{\text{env}}$ & $\mathcal{L}_{\text{homo}}$ & $\mathcal{L}_{\text{con}}$ & \multicolumn{4}{c|}{MAE $\downarrow$} & \multicolumn{4}{c}{GM $\downarrow$} \\  \midrule
&  &  & All & Many & Med & Few & All & Many & Med & Few \\ \midrule
 &  &  & 7.67 & 6.66 & 9.30 & 12.61 & 4.85 & 4.17 & 6.51 & 8.98\\
 & \checkmark & & 7.87 & 7.01 & 8.99 & 12.90 & 5.12 & 4.56 & 6.11 & 9.39\\ 
 \checkmark &  & & 7.52 & \textbf{6.63} & 8.69 & 12.63 & 4.85 & 4.27 & 5.90 & 9.48\\ 
 \checkmark & \checkmark & & 7.50 & 6.73 & 8.53 & 11.92 & 4.81 & 4.37 & 5.49 & 8.29\\
  &  & \checkmark & 7.55 & 6.73 & 8.47 & 12.71 & 4.79 & 4.24 & 5.68 & 9.42 \\
 \checkmark & \checkmark & \checkmark & \textbf{7.22} & 7.38 & \textbf{6.64} & \textbf{8.28} & \textbf{4.50} & \textbf{4.12} & \textbf{5.37} & \textbf{6.29} \\
\toprule
\end{tabular}

\label{tab:table7}
\end{table}

\subsection{Computational cost} 

In this subsection, we compare the time consumption of the Surrogate-driven Representation Learning (SRL) framework with other baseline methods for age estimation and text similarity regression tasks. The reported time consumption, expressed in seconds, represents the average training time per mini-batch update. All experiments were conducted using a GTX 3090 GPU.

Table \ref{tab:training_time} shows that SRL achieves a considerably lower training time compared to the LDS + FDS, while remaining competitive with RankSim, Balanced MSE, and Ordinal Entropy. This demonstrates SRL's ability to handle complex tasks efficiently without introducing substantial computational overhead.

\begin{table}[h]
    \centering
    \scriptsize
    \caption{Average training time per mini-batch update (in seconds) for age estimation (AgeDB-DIR) and text similarity regression (STS-B-DIR) tasks, using a GTX 3090 GPU.}
    \label{tab:training_time}
    \begin{tabular}{lcc}
        \toprule
        Method& AgeDB-DIR (s)& STS-B-DIR (s)\\
        \midrule
        VANILLA & 12.24 & 25.13 \\
        LDS + FDS & 38.42 & 44.45 \\
        RankSim & 16.86 & 30.04 \\
        Balanced MSE & 16.21 & 28.12 \\
        Ordinal Entropy & 17.29 & 29.37 \\
        SRL (Ours) & 17.10 & 27.35 \\
        \bottomrule
    \end{tabular}
\end{table}

\subsection{Impact of Bin Numbers}
\label{bin_numbers}
In our geometric framework, we employ piecewise linear interpolation to approximate the continuous path $l$. The granularity of this approximation is determined by the number of bins used for discretization, where finer binning naturally leads to smoother interpolation. To empirically analyze the impact of bin numbers ($B$) on model performance, we conducted extensive experiments across both synthetic and real-world datasets.
For the synthetic OL-DIR dataset and the real-world AgeDB-DIR dataset, we varied the number of bins across the label space. Note that for AgeDB-DIR, the finest possible bin size is constrained to 1 due to the discrete nature of age labels, while OL-DIR allows for arbitrary bin sizes. The results are presented in Table \ref{tab:bin_ablation}.

\begin{table}[htbp]
\centering
\scriptsize
\caption{Impact of bin numbers on model performance}
\label{tab:bin_ablation}
\setlength{\tabcolsep}{3pt}
\renewcommand{\arraystretch}{1.1}
\begin{tabular}{l|ccccccc}
\toprule
$B$ & 10 & 20 & 50 & 100 & 1000 & 2000 & 4000 \\
\midrule
OL-DIR (MAE $\times 10^{-3}$) & 9.92 & 9.29 & 9.20 & 9.18 & 9.18 & 9.17 & 9.18 \\
AgeDB-DIR (MAE) & 7.44 & 7.38 & 7.31 & 7.22 & - & - & -  \\ 
\bottomrule
\end{tabular}
\end{table}

\subsection{Experiments on UCI-DIR (Table \ref{tab:airfoil_complete}, \ref{tab:abalone_complete}, \ref{tab:realestate_complete}, \ref{tab:concrete_complete})} 
\label{uci_eval}

\begin{table}[htbp]
\centering
\scriptsize

\caption{Complete results on UCI-DIR for Airfoil (MAE with standard deviation), the best results are \textbf{bold}.}
\setlength{\tabcolsep}{3pt}
\renewcommand{\arraystretch}{1.1}
\begin{tabular}{l|cccc}
\toprule
Metrics & \multicolumn{4}{c}{MAE}  \\ 
\midrule
Shot & All & Many & Med & Few \\ 
\midrule
VANILLA          & 5.657(0.324) & 5.112(0.207) & 5.031(0.445) & 6.754(0.423) \\
LDS + FDS        & 5.761(0.331) & \textbf{4.445}(0.208) & 4.792(0.412) & 7.792(0.499) \\
RankSim          & 5.228(0.335) & 5.049(0.92) & 4.908(0.786) & 5.718(0.712) \\
BalancedMSE      & 5.694(0.342)& 4.512(0.179)& 5.035(0.554)& 7.277(0.899)\\
Ordinal Entropy  & 6.270(0.415)& 4.847(0.223)& 5.369(0.635)& 8.315(0.795)\\ 
SRL (ours)       & \textbf{5.100}(0.286)& 4.832(0.098)& \textbf{4.745}(0.336)& \textbf{5.693}(0.542)\\
\toprule
\end{tabular}
\label{tab:airfoil_complete}
\end{table}

\begin{table}[htbp]
\centering
\scriptsize

\caption{Complete results on UCI-DIR for Abalone (MAE with standard deviation), the best results are \textbf{bold}.}
\setlength{\tabcolsep}{3pt}
\renewcommand{\arraystretch}{1.1}
\begin{tabular}{l|cccc}
\toprule
Metrics & \multicolumn{4}{c}{MAE} \\ 
\midrule
Shot & All & Many & Med & Few  \\ 
\midrule
VANILLA          & 4.567(0.211)& \textbf{0.878}(0.152)& 2.646(0.349)& 7.967(0.344)\\
LDS + FDS        & 5.087(0.456)& 0.904(0.245)& 3.261(0.435)& 9.261(0.807)\\
RankSim          & 4.332(0.403)& 0.975(0.067)& 2.591(0.516)& 7.421(0.966)\\
BalancedMSE      & 5.366(0.542)& 2.135(0.335)& 2.659(0.456)& 9.368(0.896)\\
Ordinal Entropy  & 6.774(0.657)& 2.314(0.256)& 4.013(0.654)& 11.610(1.275)\\ 
SRL (ours)       & \textbf{4.158}(0.196)& 0.892(0.042)& \textbf{2.423}(0.199)& \textbf{7.191}(0.301)\\
\toprule
\end{tabular}
\label{tab:abalone_complete}
\end{table}

\begin{table}[htbp]
\centering
\scriptsize

\caption{Complete results on UCI-DIR for Real Estate (MAE with standard deviation), the best results are \textbf{bold}.}
\setlength{\tabcolsep}{3pt}
\renewcommand{\arraystretch}{1.1}
\begin{tabular}{l|cccc}
\toprule
Datasets & \multicolumn{4}{c}{MAE} \\ 
\midrule
Shot & All & Many & Med & Few   \\ 
\midrule
VANILLA          & 0.326(0.003) & 0.273(0.005) & 0.376(0.003) & 0.365(0.012) \\
LDS + FDS        & 0.346(0.004) & 0.325(0.002) & 0.400(0.002) & 0.335(0.023) \\
RankSim          & 0.373(0.008) & 0.343(0.004) & 0.381(0.008) & 0.397(0.032) \\
BalancedMSE      & 0.337(0.007) & 0.313(0.004) & 0.398(0.009) & 0.326(0.028) \\
Ordinal Entropy  & 0.339(0.007) & 0.286(0.004) & 0.421(0.005) & 0.351(0.031) \\ 
SRL (ours)       & \textbf{0.278}(0.002) & \textbf{0.262}(0.006) & \textbf{0.296}(0.005) & \textbf{0.287}(0.023) \\
\toprule
\end{tabular}
\label{tab:realestate_complete}
\end{table}

\begin{table}[htbp]
\centering
\scriptsize

\caption{Complete results on UCI-DIR for Concrete (MAE with standard deviation), the best results are \textbf{bold}.}
\setlength{\tabcolsep}{3pt}
\renewcommand{\arraystretch}{1.1}
\begin{tabular}{l|cccc}
\toprule
Datasets & \multicolumn{4}{c}{MAE}\\ 
\midrule
Shot & All & Many & Med & Few \\ 
\midrule
VANILLA          & 7.287(0.364) & 5.774(0.289) & 6.918(0.346) & 9.739(0.487) \\
LDS + FDS        & 6.879(0.344) & 6.210(0.310) & 6.730(0.337) & 7.594(0.380) \\
RankSim          & 6.714(0.336) & 5.996(0.300) & 5.574(0.279) & 9.456(0.473) \\
BalancedMSE      & 7.033(0.352) & \textbf{4.670}(0.234) & 6.368(0.318) & 9.722(0.486) \\
Ordinal Entropy  & 7.115(0.356) & 5.502(0.275) & 6.358(0.318) & 9.313(0.466) \\ 
SRL (ours)       & \textbf{5.939}(0.297) & 5.318(0.266) & \textbf{5.800}(0.290) & \textbf{6.603}(0.330) \\
\toprule
\end{tabular}
\label{tab:concrete_complete}
\end{table}

\newpage

\subsection{Experiments on AgeDB-DIR}
\label{agedb_app}
\textbf{Training Details:} In Table \ref{tab:complete_agedb}, our primary results on AgeDB-DIR encompasses the replication of all baseline models on an identical server configuration (RTX 3090), adhering to the original codebases and training recipes. We observe a performance drop in RankSim \citep{gong2022ranksim} and ConR \citep{keramati2023conr} in comparison to the results reported in their respective studies. To ensure a fair comparison, we present the \textbf{mean and standard deviation (in parentheses)} of the performances for SRL (ours), RankSim, and ConR, based on three independent runs. We found SRL superiors performance in most categories and all Med-shot and Few-shot metrics.

We would like to note that we found self-conflict performance in the original ConR \cite{keramati2023conr} paper, where they report overall MAE of 7.20 in main result (Table 1) and 7.48 in the ablation studies (Table 6). \textbf{The results in Table 6 are closed to our reported result}.

\begin{table*}[htbp]
\centering
\caption{Complete Results on AgeDB-DIR}
\setlength{\tabcolsep}{3pt}
\renewcommand{\arraystretch}{1.1}
\resizebox{1.55\columnwidth}{!}{
\begin{tabular}{l|l|ccccccc}
\toprule
Metrics & Shot & VANILLA & LDS + FDS & RankSim & BalancedMSE & Ordinal Entropy & ConR & \textbf{SRL (ours)} \\ 
\midrule
\multirow{4}{*}{MAE$\downarrow$} & All & 7.67 & 7.55 & 7.41{(0.03)} & 7.98 & 7.60 & 7.41{(0.02)} & \textbf{7.22}{(0.02)} \\
                     & Many & 6.66 & 7.03 & \textbf{6.49}{(0.01)} & 7.58 & 6.69 & 6.51{(0.02)} & 6.64{(0.01)} \\
                     & Med & 9.30 & 8.46 & 8.73(0.05) & 8.65 & 8.87 & 8.81(0.03) & \textbf{8.28}{(0.04)} \\
                     & Few & 12.61 & 10.52 & 12.47(0.09) & 9.93 & 12.68 & 12.04(0.04) & \textbf{9.81}{(0.05)} \\ \midrule
\multirow{4}{*}{GM$\downarrow$}  & All & 4.85 & 4.86 & 4.71{(0.03)} & 5.01 & 4.91 & 4.70{(0.02)} & \textbf{4.50}{(0.02)} \\
                     & Many & 4.17 & 4.57 & 4.15(0.02) & 4.83 & 4.28 & 4.13(0.02) & \textbf{4.12}{(0.02)} \\
                     & Med & 6.51 & 5.38 & 5.74(0.04) & 5.46 & 6.20 & 5.91(0.06) & \textbf{5.37}{(0.02)} \\
                     & Few & 8.98 & 6.75 & 8.92(0.08) & 6.30 & 9.29 & 8.59(0.0) & \textbf{6.29}{(0.04)} \\ \midrule
\multirow{4}{*}{MSE$\downarrow$} & All & 100.01 & 97.05 & 94.37(0.10) & 107.35 & 97.28 & 92.57{(0.06)} & \textbf{91.71}{(0.02)} \\
                     & Many & 76.67 & 82.68 & \textbf{72.00}(0.09) & 95.49 & 74.79 & 72.06{(0.04)} & 77.23{(0.05)} \\
                     & Med & 130.21 & 114.00 & 121.38(2.15) & 125.55 & 122.07 & 121.24{(1.88)} & \textbf{115.65}{(1.42)} \\
                     & Few & 237.00 & 185.98 & 230.97(3.22) & 169.00 & 241.13 & 207.00{(3.09)} & \textbf{162.22}{(2.08)} \\
\toprule
\end{tabular}}
\label{tab:complete_agedb}
\end{table*}
\begin{table*}[htbp]
\centering
\caption{Complete Results on IMDB-WIKI-DIR}
\setlength{\tabcolsep}{3pt}
\renewcommand{\arraystretch}{1.1}
\resizebox{1.55\columnwidth}{!}{\begin{tabular}{l|c|ccccccc}
\toprule
Metrics & Shot & VANILLA & LDS + FDS & RankSim & BalancedMSE & Ordinal Entropy & ConR & SRL (ours) \\ 
\midrule
\multirow{4}{*}{MAE$\downarrow$} & All & 8.03 & 7.73 & 7.72 & 8.43 & 8.01 & 7.84(0.04) & \textbf{7.69}(0.02) \\
                     & Many & 7.16 & 7.22 & \textbf{6.92} & 7.84 & 7.17 & 7.15(0.03) & 7.08(0.01) \\
                     & Med & 15.48 & 12.98 & 14.52 & 13.35 & 15.15 & 14.36(0.04) & \textbf{12.65}(0.04) \\
                     & Few & 26.11 & 23.71 & 25.89 & 23.27 & 26.48 & 25.15(0.06) & \textbf{22.78}(0.06) \\ \midrule
\multirow{4}{*}{GM$\downarrow$}  & All & 4.54 & 4.40 & 4.29 & 4.93 & 4.47 & 4.43(0.04) & \textbf{4.28}(0.02) \\
                     & Many & 4.14 & 4.17 & 3.92 & 4.68 & 4.07 & 4.05(0.03) & \textbf{4.03}(0.02) \\
                     & Med & 10.84 & 7.87 & 9.72 & 7.90 & 10.56 & 9.91(0.05) & \textbf{7.28}(0.03) \\
                     & Few & 18.64 & 15.77 & 18.02 & 15.51 & 21.11 & 18.55(0.06) & \textbf{15.25}(0.05) \\ \midrule
\multirow{4}{*}{MSE$\downarrow$} & All & 136.04 & 130.56 & 130.95 & 146.19 & 137.50 & 132.41(1.22) & \textbf{129.97}(0.93) \\
                     & Many & 105.72 & 106.93 & \textbf{102.06} & 121.64 & 107.62 & 105.29(0.88) & 105.83(0.77) \\
                     & Med & 373.07 & 315.92 & 351.22 & 343.12 & 369.88 & 338.30(1.99) & \textbf{311.17}(1.25) \\
                     & Few & 978.00 & 861.15 & 977.82 & 787.71 & 976.56 & 934.12(3.03) & \textbf{859.81}(2.28) \\
\toprule
\end{tabular}}
\label{tab:complete_imdb}
\end{table*}
\begin{table*}[htbp]
\centering
\caption{Complete Results on STS-B-DIR}
\setlength{\tabcolsep}{3pt}
\renewcommand{\arraystretch}{1.1}
\resizebox{1.55\columnwidth}{!}{\begin{tabular}{l|l|cccccc}
\toprule
Metrics & Shot & VANILLA & LDS + FDS & RankSim & BalancedMSE & Ordinal Entropy & SRL (ours)\\ 
\midrule
\multirow{4}{*}{MSE $\downarrow$}
& All & 0.993 & 0.900 & 0.889 & 0.909 & 0.943 & \textbf{0.877} \\
& Many & 0.963 & 0.911 & 0.907 & 0.894 & 0.902 & \textbf{0.886} \\
& Med & 1.000 & 0.881 & 0.874 & 1.004 & 1.161 & \textbf{0.873} \\
& Few & 1.075 & 0.905 & 0.757 & 0.809 & 0.812 & \textbf{0.745} \\ \midrule
\multirow{4}{*}{Pearson correlation $\uparrow$}
& All & 0.742 & 0.757 & 0.763 & 0.757 & 0.750 & \textbf{0.765} \\
& Many & 0.685 & 0.698 & 0.708 & 0.703 & 0.702 & \textbf{0.708} \\
& Med & 0.693 & 0.723 & 0.692 & 0.685 & 0.679 & \textbf{0.749} \\
& Few & 0.793 & 0.806 & 0.842 & 0.831 & 0.767 & \textbf{0.844} \\ \midrule
\multirow{4}{*}{MAE $\downarrow$}
& All & 0.804 & 0.768 & 0.765 & 0.776 & 0.782 & \textbf{0.750} \\
& Many & 0.788 & 0.772 & 0.772 & 0.763 & 0.756 & \textbf{0.748} \\
& Med & 0.865 & 0.785 & 0.779 & 0.839 & 0.900 & \textbf{0.773} \\
& Few & 0.837 & 0.712 & 0.699 & 0.749 & 0.762 & \textbf{0.694} \\ \midrule
\multirow{4}{*}{Spearman correlation $\uparrow$}
& All & 0.740 & 0.760 & 0.767 & 0.762 & 0.755 & \textbf{0.769} \\
& Many & 0.650 & 0.670 & 0.685 & 0.677 & 0.669 & \textbf{0.689} \\
& Med & 0.495 & 0.488 & 0.495 & 0.487 & 0.448 & \textbf{0.503} \\
& Few & 0.843 & 0.819 & 0.862 & 0.867 & 0.845 & \textbf{0.879} \\
\toprule
\end{tabular}}
\label{tab:complete_stsb}
\end{table*}

\begin{table*}[htbp]
\centering
\caption{Complete results on OL-DIR with standard deviation added, best results are \textbf{bold}.}
\setlength{\tabcolsep}{3pt} 
\renewcommand{\arraystretch}{1.1} 
\resizebox{1.55\columnwidth}{!}{\begin{tabular}{l|cccc|cccc}
\toprule
Operation & \multicolumn{4}{c|}{ MAE($10^{-3}$)  $\downarrow$} & \multicolumn{4}{c}{ MSE ($10^{-4}$) $\downarrow$} \\ \midrule
Shot & All & Many & Med & Few & All & Many & Med & Few \\ \midrule
\textit{\textbf{Linear}} \\ \midrule
VANILLA & 15.64(2.72) &  11.86(2.20) & 15.45(3.55) & 27.00(5.62) & 5.40(1.10) & 2.81(0.75) & 4.40(1.23) & 14.20(2.25) \\
Ordinal Entropy & 10.07(1.22)  & 9.26(0.98) & 9.85(1.45) & 13.01(1.92) & 2.00(0.32) & 1.53(0.19) & 1.89(0.73) & 3.42(0.82) \\ 
SRL (ours)  & \textbf{9.18}(0.92) & \textbf{8.32}(0.66) & \textbf{9.47}(1.13) & \textbf{9.33}(1.89)  & \textbf{1.98}(0.37) & \textbf{0.98}(0.21) & \textbf{1.72}(0.62) & \textbf{2.67}(0.99) \\ \midrule
\textit{\textbf{Nonlinear}} \\ \midrule
VANILLA  & 11.64(1.87) & 9.89(1.25) & 11.02(2.23) & 19.77(2.89) & 9.20(1.23) & \textbf{4.33}(0.88) & 7.53(1.55) & 24.70(1.99) \\
Ordinal Entropy  & 12.91(1.25) & 9.93(0.93) & 13.07(1.57) & 21.02(1.89) & 13.80(2.98) & 8.82(2.25) & 11.84(3.59) & 30.12(5.40) \\
SRL (ours)  & \textbf{11.25}(1.13) & \textbf{9.48}(0.75) & \textbf{9.22}(1.45) &  \textbf{17.00}(1.54) &  \textbf{8.60}(1.04) & {7.42}(0.70) & \textbf{6.41}(1.15) & \textbf{14.12}(1.39) \\
\toprule
\end{tabular}}
\label{tab:complete_iol}
\end{table*}

\subsection{Experiment on IMDB-WIKI-DIR}
\label{imdb}

\textbf{Training Details:} In Table \ref{tab:complete_imdb}, our primary results on IMDB-WIKI-DIR encompass the replication of all baseline models on an identical server configuration (RTX 3090), adhering to the original codebases and training receipes. We observe a performance drop of ConR \citep{keramati2023conr} in comparison to the results reported in their respective studies. To ensure a fair comparison, we present the \textbf{mean and standard deviation (in parentheses)} of the performances for SRL (ours) and ConR, based on three independent runs. We found SRL superiors performance in most categories and all Med-shot and Few-shot metrics.

We would like to note that we found self-conflict performance in the original ConR \cite{keramati2023conr} paper, where they report overall MAE of 7.33 in the main result (Table 2) and 7.84 in the ablation studies (Table 8), \textbf{The results in Table 8 are close to our reported result}.

\subsection{Complete result on STS-B-DIR (Table \ref{tab:complete_stsb})}
\label{stsb}

\subsection{Complete result on Operator Learning ( Table \ref{tab:complete_iol})}
\label{ol_result}

\section{Pseudo Code (Algorithm \ref{pseudo_code}) for Surrogate-driven Representation Learning (SRL)}
\label{b}

\begin{algorithm*}

\caption{Pseudo Code for Surrogate-driven Representation Learning (SRL)}
\begin{algorithmic}[1]
\REQUIRE Training set \(D = \{(x_i, y_i)\}_{i=1}^N\),  encoder \(f\), regression function \(g\), total training epochs \(E\),  momentum \(\alpha\), a set of uniformly distributed points ${U}$, surrogate ${S}$, batch size ${M}$.
\FOR{\(e = 0\) to \(E\)}
    \REPEAT
        \STATE Sample a mini-batch \(\{(x_m, y_m)\}_{m=1}^M\) from \(D\)
        \STATE   \(\{z_m\}_{m=1}^M = f(\{x_m\}_{m=1}^M)\)
        \IF{\(e=0\)}
            \STATE Update the model with loss \(\mathcal{L}=\mathcal{L}_{reg}(\{y_m\}_{m=1}^M,g(\{z_m\}_{m=1}^M))\)
        \ELSE
        \STATE get \(C\) from \(\{z_m\}_{m=1}^M\) using Equation (8)
        \STATE get \(S^{e'} \) from \(C\) and \(S^e\) using Equation (9)
        \STATE Update the model with loss \(\mathcal{L}=\mathcal{L}_{reg}(\{y_m\}_{m=1}^M,g(\{z_m\}_{m=1}^m))+\mathcal{L}_{G}(S^{e'}, U)+\mathcal{L}_{con}(S^{e'},\{z_m\}_{m=1}^M)\)

        \ENDIF
    \UNTIL iterate over all training samples at current epoch \(e\)
    \STATE // Update the surrogate
    \STATE get \(\hat{S}^e\)by calculate the class center for the current epoch
    \IF{\(e=0\)}
    \STATE \(S^{1}=\hat{S^{e}}\)
    \ELSE
    \STATE \(S^{e+1}=\alpha S^{e}+(1-\alpha) \hat{S}^e\) \# Momentum update the surrogate, Equation (9)
    \ENDIF
\ENDFOR
\end{algorithmic}
\label{pseudo_code}
\end{algorithm*}

\section {Broader impacts}
\label{E}

We introduce novel geometric constraints to the representation learning of imbalanced regression, which we believe will significantly benefit regression tasks across various real-world applications. Currently, we are not aware of any potential negative societal impacts.

\section{Limitation and Future Direction}
\label{F}

In considering the limitations and future directions of our research, it's important to acknowledge that our current methodology has not delved into optimizing the feature distribution in scenarios involving regression with higher-dimensional labels. This presents a notable area for future exploration. Additionally, investigating methods to effectively handle complex label structures in imbalanced regression scenarios could significantly enhance the applicability and robustness of our proposed techniques.